# Properties of Bethe Free Energies and Message Passing in Gaussian Models

**Botond Cseke** B.CSEKE@SCIENCE.RU.NL
**Tom Heskes** T.HESKES@SCIENCE.RU.NL
*Institute for Computing and Information Sciences*
*Faculty of Science, Radboud University Nijmegen*
*Heyendaalseweg 135, 6525 AJ, The Netherlands*

## Abstract

We address the problem of computing approximate marginals in Gaussian probabilistic models by using mean field and fractional Bethe approximations. We define the Gaussian fractional Bethe free energy in terms of the moment parameters of the approximate marginals, derive a lower and an upper bound on the fractional Bethe free energy and establish a necessary condition for the lower bound to be bounded from below. It turns out that the condition is identical to the pairwise normalizability condition, which is known to be a sufficient condition for the convergence of the message passing algorithm. We show that stable fixed points of the Gaussian message passing algorithm are local minima of the Gaussian Bethe free energy. By a counterexample, we disprove the conjecture stating that the unboundedness of the free energy implies the divergence of the message passing algorithm.

## 1. Introduction

One of the major tasks of probabilistic inference is calculating marginal posterior probabilities of a set of variables given some observations. In case of Gaussian models, the computational complexity of computing marginals might scale cubically with the number of variables, while for models with discrete variables it often leads to intractable computations. Computations can be made faster or tractable by using approximate inference methods like the mean field approximation (e.g., Jaakkola, 2000) and the Bethe-type approximation (e.g., Yedidia, Freeman, & Weiss, 2000). These methods were developed for discrete probabilistic graphical models, but they are applicable to Gaussian models as well. However, there are important differences in their behavior for the discrete and Gaussian cases. For example, while in discrete models the error function of the Bethe approximation—called Bethe free energy—is bounded from below (Heskes, 2004; Watanabe & Fukumizu, 2009), in Gaussian models this might not always the case (Welling & Teh, 2001).

An understanding of properties of the Bethe free energy of Gaussian models might also be help to understand the properties of the energy function in conditional Gaussian models. Conditional Gaussian or hybrid graphical models, such as switching Kalman filters (Zoeter & Heskes, 2005), combine both discrete and Gaussian variables. Approximate inference in these models can be carried out by expectation propagation (e.g., Minka, 2004, 2005) which can be viewed as a generalization of the Bethe approximation, where the marginal consistency constraints on the approximate marginals are replaced by expectation constraints (Heskes, Opper, Wiegerinck, Winther, & Zoeter, 2005). In order to understand the properties of the Bethe free energy of hybrid models, a good understanding of the two special cases of discrete and Gaussian models is needed. While the properties of the Bethe free energy of discrete models have been studied extensively in the last decade and are well understood (Yedidia et al., 2000; Heskes, 2003; Wainwright, Jaakkola, & Willsky, 2003; Watanabe & Fukumizu, 2009), the properties of the Gaussian Bethe free energy have been studied much less.

The message passing algorithm is a well established method for finding the stationary points of the Bethe free energy (Yedidia et al., 2000; Heskes, 2003). It works by locally updating the approximate marginals and has been successfully applied in both discrete (e.g., Murphy, Weiss, & Jordan, 1999; Wainwright et al., 2003) and Gaussian models (e.g., Weiss & Freeman, 2001; Rusmevichientong & Roy, 2001; Malioutov, Johnson, & Willsky, 2006; Johnson, Bickson, & Dolev, 2009; Nishiyama & Watanabe, 2009; Bickson, 2009). Gaussian message passing is the simplest case of a free-energy based message passing algorithm on models with continuous variables, therefore, it is important to understand its behavior.

Gaussian message passing has many practical applications like in distributed averaging (Moallemi & Roy, 2006), peer-to-peer rating, linear detection, SVM regression (Bickson, 2009) and more generally in problems that involve solving large sparse linear systems or approximating the marginal variances of large sparse Gaussian systems typically encountered in distributed computing settings. For further applications the reader is referred to the work of Bickson (2009) and references therein.

Finding sufficient conditions for the convergence of message passing in Gaussian models has been successfully addressed by many authors. Using the computation tree approach, Weiss and Freeman (2001) proved that message passing converges whenever the precision matrix—inverse covariance—of the probability distribution is diagonally dominant[1]. With the help of an analogy between message passing and walk–sum analysis, (Malioutov et al., 2006) derived the stronger condition of pairwise normalizability[2]. A different approach was taken by Welling and Teh (2001), who directly minimized the Bethe free energy with regard to the parameters of approximate marginals, conjecturing that Gaussian message passing converges if and only if the free energy is bounded from below. Their experiments showed that message passing and direct minimization either converge to the same solution or both fail to converge. We adopt a similar approach, that is, instead of analyzing the properties of the Gaussian message passing algorithm using approaches like in Weiss and Freeman or Malioutov et al., we choose to study the properties of the Gaussian Bethe free energy and its stationary points. This will help us to draw conclusions about the existence of local minima, the possible stable fixed points to which message passing can converge.

This paper is structured as follows. In Section 2 we introduce Gaussian Markov random fields and the message passing algorithm. In Section 3 we define the Gaussian fractional Bethe free energies parameterized by the moment parameters of the approximate marginals and derive boundedness conditions for them. These two sections are based on the authors earlier work (Cseke & Heskes, 2008). In Section 4 we analyze the stability properties of the Gaussian message passing algorithm and, using a similar line of argument as Watanabe and

1. The matrix $\boldsymbol{A}$ is diagonally dominant if $|A_{ii}| > \sum_{j \neq i} |A_{ij}|$ for all $i$.
2. Following the work of Malioutov et al. (2006), we call a Gaussian distribution pairwise normalizable if it can be factorized into a product of normalizable "pair" factors, that is, $p(x_1, \ldots, x_n) = \prod_{ij} \Psi_{ij}(x_i, x_j)$ such that all $\Psi_{ij}$s are normalizable.

Fukumizu (2009), we show that its stable fixed points are indeed local minima of the Bethe free energy. We conclude the paper with a few experiments in Sections 5 and 6 supporting our results and their implications.

## 2. Approximating Marginals in Gaussian Models

The probability density of a Gaussian random vector $\boldsymbol{x} \in \mathbb{R}^n$ is defined in terms of canonical parameters $\boldsymbol{h}$ and $\boldsymbol{Q}$ as

$$p(\boldsymbol{x}) \propto \exp\left\{\boldsymbol{h}^T\boldsymbol{x} - \frac{1}{2}\boldsymbol{x}^T\boldsymbol{Q}\boldsymbol{x}\right\}, \tag{1}$$

where $\boldsymbol{Q}$ is s positive definite matrix. The expectation $\boldsymbol{m}$ and the covariance $\boldsymbol{V}$ of $\boldsymbol{x}$ is then given by $\boldsymbol{m} = \boldsymbol{Q}^{-1}\boldsymbol{h}$ and $\boldsymbol{V} = \boldsymbol{Q}^{-1}$ respectively. In many real world applications the matrix $\boldsymbol{Q}$ is sparse with and it typically has low density, that is, the number of non-zero elements in $\boldsymbol{Q}$ scales with the number of variables $n$.

This probability density can also be defined in terms of an undirected probabilistic graphical model commonly known as Gaussian Markov random field (GMRF). Since the interactions between the variables in $p$ are pairwise, we can associate the variables $x_i$ to the nodes $v \in V = \{1, \ldots, n\}$ of an undirected graph $G = (V, E)$, where the edges $e \in E \subseteq V \times V$ of the graph stand for the non-zero off-diagonal elements of $\boldsymbol{Q}$. We use $i \sim j$ as a proxy for $(i, j) \in E$. By using the notation introduced above, the density $p$ in (1) can be written as the product

$$p(\boldsymbol{x}) \propto \prod_{i \sim j} \Psi_{ij}(x_i, x_j) \tag{2}$$

of Gaussian functions $\Psi_{ij}(x_i, x_j)$ (also called potentials) associated with the edges $e = (i, j)$ of the graph. If $\boldsymbol{h}$ and $\boldsymbol{Q}$ are given then we can define the potentials as

$$\Psi_{ij}(x_i, x_j) = \exp\left\{\gamma_{ij}^i h_i x_i + \gamma_{ij}^j h_j x_j - \gamma_{ij}^i Q_{ii} x_i^2/2 - \gamma_{ij}^j Q_{jj} x_j^2/2 - Q_{ij} x_i x_j\right\},$$

where $\sum_{i \sim j} \gamma_{ij}^i = 1$ and $\sum_{j \sim i} \gamma_{ij}^j = 1$ are partitioning $\boldsymbol{h}$ and $\boldsymbol{Q}$ into the corresponding factors. In practice, however, the factors $\Psi_{ij}$ might be given by the problem at hand and $\boldsymbol{h}$ and $\boldsymbol{Q}$ as well as $\gamma_{ij}^i$ and $\gamma_{ij}^j$ computed by summing their parameters and computing the partitioning respectively. Without loss of generality, we can and we will use $Q_{ii} = 1$, since the results in the paper can be easily re-formulated for general $\boldsymbol{Q}$s by a rescaling of the variables (e.g., Malioutov et al., 2006).

The numerical calculation of all marginals, can be done by solving the linear system $\boldsymbol{m} = \boldsymbol{Q}^{-1}\boldsymbol{h}$ and performing a sparse Cholesky factorization $\boldsymbol{L}\boldsymbol{L}^T = \boldsymbol{Q}$ followed by solving the Takahashi equations (Takahashi, Fagan, & Chin, 1973). An alternative option to calculate the marginal means and to *approximate marginal variances* is to run the Gaussian message passing algorithm in the probabilistic graphical model associated with the representation in (2). The Gaussian message passing algorithm is the Gaussian variant of message passing algorithm (Pearl, 1988), which is a dynamical programming algorithm introduced to compute marginal densities in discrete probabilistic models with pairwise interactions and tree-structured graphs $G$. However, it turned out that by running it in loops on graphs with cycles, it yields good approximations of the marginal distributions (Murphy et al., 1999). Weiss and Freeman (2001) showed that when the Gaussian message passing algorithm is converging, it computes the exact mean parameters $\boldsymbol{m}$, thus it can also be used for solving linear systems (e.g., Bickson, 2009). Message passing works by updating and passing directed messages along the edges of the graph $G$, which, in case the algorithm converges, are then used to compute (approximate) marginal probability distributions. The Gaussian and the discrete algorithms have the same functional form with the exception of the summation (discrete case) and integration operators (Gaussian case). Each message $\mu_{i\leftarrow j}(x_i)$ is updated according to

$$\mu^{new}_{i\leftarrow j}(x_i) = \int dx_j \, \Psi_{ij}(x_i, x_j) \prod_{k\in\partial j\setminus i} \mu_{j\leftarrow k}\left(x_j\right), \tag{3}$$

where $\partial i = \{j : j \sim i\}$ denotes the index set of variables connected to $x_i$ in $G$. At each step the current approximations $q_{ij}(x_{i,j})$ of $p(x_i, x_j)$ can be computed according to

$$q_{ij}\left(x_i, x_j\right) \propto \Psi_{ij}(x_i, x_j) \prod_{l\in\partial i\setminus j} \mu_{i\leftarrow l}\left(x_i\right) \prod_{k\in\partial j\setminus i} \mu_{j\leftarrow k}\left(x_j\right). \tag{4}$$

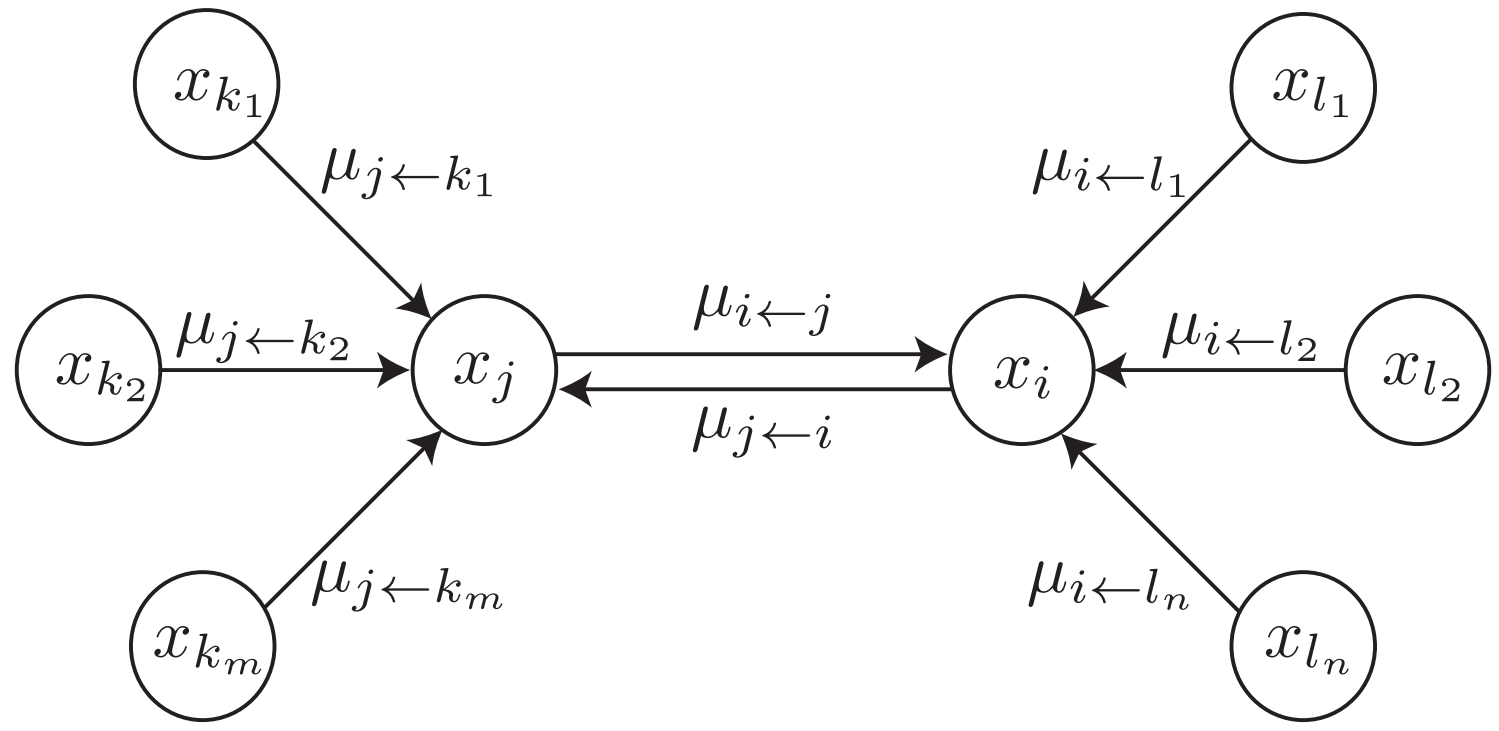

Figure 1: An illustration of the incoming and outgoing messages at adjacent nodes $i$ and $j$.

The update steps in (9) have to be iterated until convergence. The corresponding $q_{ij}(x_i, x_j)$s yield the final approximation of the $p(x_i, x_j)$s. It is common to use damping, that is, to replace $\mu^{new}_{i\leftarrow j}(x_i)$ by $\mu_{i\leftarrow j}(x_i)^{1-\epsilon}\mu^{new}_{i\leftarrow j}(x_i)^{\epsilon}$ with $\epsilon \in (0, 1]$. In practice, this helps to dampen the possible periodic paths of (3), but it keeps the properties of the fixed points unchanged. Figure 1 illustrates the incoming and outgoing messages at the nodes associated with variables $x_i$ and $x_j$. A quite significant difference between the discrete and Gaussian the message passing is the replacement of the sum operator with the integral operator. While finite sums always exist, the integral in (3) can become infinite. This problem can be remedied technically by a canonical parameterization (see Section 4) which keeps the algorithm running, but it can lead to non-normalizable approximate marginals $q_{ij}$, and thus a (possible) break-down of the algorithm.

Message passing was introduced by Pearl (1988) as a heuristic algorithm (in discrete models), however, Yedidia et al. (2000) showed that it can also be viewed as an algorithm for finding the stationary points of the so-called Bethe free energy, an error function measuring the difference between $p$ and a specific family of distributions to be detailed in the next section. It has been shown by Heskes (2003) and later in a different way by Watanabe and Fukumizu (2009) that stable fixed points of the (loopy) message passing algorithm are local minima of the corresponding Bethe free energy. In this paper we show that this holds for Gaussian models as well.

Our interest in the properties of the Gaussian Bethe free energy and the corresponding Gaussian message passing algorithm is motivated mainly by their implications in more general models and inference algorithms like non-Gaussian models and expectation propagation, respectively. For this reason, we will not compare the speed of the method and the accuracy of the approximation with the above mentioned exact linear algebraic methods.

As mentioned in the introduction, the approach we take is similar to that of Welling and Teh (2001), that is, we study the properties of the Gaussian Bethe free energy, parameterized in terms of the moment parameters of the approximate marginals. In the following we introduce the mean field and the Bethe approximation in Gaussian models. Readers familiar with this subject can continue with Section 3.

### 2.1 The Gaussian Bethe Free Energy

A popular method to approximate marginals is approximating $p$ with a distribution $q$ having a form that makes marginals easy to identify, for example, it factorizes or it has a "tree-like" form. The most common quantity to measure the difference between two probability distributions is the Kullback-Leibler divergence $\mathrm{D}\left[q\,||\,p\right]$. It is often used to characterize the quality of the approximation and formulate the computation of approximate marginals as the optimization problem

$$q^*(\boldsymbol{x}) = \underset{q\in\mathcal{F}}{\operatorname{argmin}} \int d\boldsymbol{x}\, q(\boldsymbol{x}) \log\left[\frac{q(\boldsymbol{x})}{p(\boldsymbol{x})}\right]. \tag{5}$$

Here, $\mathcal{F}$ is the set of distributions with the above mentioned form. Since it is not symmetric, the Kullback-Leibler divergence is not a distance, but $\mathrm{D}\left[q\,||\,p\right] \geq 0$ for any proper $q$ and $p$, $\mathrm{D}\left[q\,||\,p\right] = 0$ if and only if $p = q$, and it is convex both in $q$ and $p$.

A family $\mathcal{F}$ of densities possessing a form that makes marginals easy to identify is the family of distributions that factorize as $q(\boldsymbol{x}) = \prod_k q_k(x_k)$. In other words, in problem (5) we approximate $p$ with a distribution that has independent variables. An approximation $q$ of this type is called mean field approximation (e.g., Jaakkola, 2000). Defining $F_{\mathrm{MF}}\left(\{q_k\}\right) = \mathrm{D}\left[\prod q_k\,||\,p\right]$ and writing out the right hand side of (5) in detail, one gets

$$F_{\mathrm{MF}}(\{q_k\}) = -\int d\boldsymbol{x} \prod_k q_k(x_k) \log p(\boldsymbol{x}) + \sum_k \int dx_k\, q_k(x_k) \log q_k(x_k).$$

Using the parameterization $q_k(x_k) = N(x_k|m_k, v_k)$, $\boldsymbol{m} = (m_1, \ldots, m_n)^T$ and $\boldsymbol{v} = (v_1, \ldots, v_n)^T$, this reduces to

$$F_{\mathrm{MF}}\left(\boldsymbol{m}, \boldsymbol{v}\right) = -\boldsymbol{h}^T\boldsymbol{m} + \frac{1}{2}\boldsymbol{m}^T\boldsymbol{Q}\boldsymbol{m} + \frac{1}{2}\sum_k Q_{kk}v_k - \frac{1}{2}\sum_k \log(v_k) + C_{\mathrm{MF}},$$

where $C_{\mathrm{MF}}$ is an irrelevant constant. Although $\mathrm{D}\left[\prod_k q_k \,||\, p\right]$ might not be convex in $(q_1, \ldots, q_n)$, one can easily check that $F_{\mathrm{MF}}$ is convex in its variables $\boldsymbol{m}$ and $\boldsymbol{v}$ and its minimum is obtained for $\boldsymbol{m} = \boldsymbol{Q}^{-1}\boldsymbol{h}$ and $v_k = 1/Q_{kk}$. Since

$$\left[\boldsymbol{Q}^{-1}\right]_{kk} = \left(Q_{kk} - \boldsymbol{Q}_{k,\backslash k}^T \left[\boldsymbol{Q}_{\backslash k,\backslash k}\right]^{-1} \boldsymbol{Q}_{\backslash k,k}\right)^{-1},$$

one can easily see that the mean field approximation underestimates variances. The mean field approximation computes a solution in which the means are exact, but the variances are computed as if there were no interactions between the variables, namely, as if the matrix $\boldsymbol{Q}$ were diagonal, thus giving poor estimates of the variances.

In order to improve the estimates for variances, one has to choose approximating distributions $q$ that are able to capture dependencies between the variables in $p$. It can be verified that any distribution in which the dependencies form a tree graph can be written in the form

$$p(\boldsymbol{x}) = \prod_{i\sim j} \frac{p(x_i, x_j)}{p(x_i)p(x_j)} \prod_k p(x_k),$$

where $i$ and $j$ run through the edges $(i, j)$ of the tree and $k$ through the nodes $1, \ldots, n$. Although in most cases the undirected graph generated by the non-zero elements in $\boldsymbol{Q}$ is not a tree, based on the "tree intuition" one can construct $q$ from one and two variable marginals as

$$q(\boldsymbol{x}) \propto \prod_{i\sim j} \frac{q_{ij}(x_i, x_j)}{q_i(x_i)q_j(x_j)} \prod_k q_k(x_k) \tag{6}$$

and constrain the functions $q_{ij}$ and $q_k$ to be marginally consistent and normalize to 1, that is, $\int dx_j q_{ij}(x_i, x_j) = q_i(x_i)$ for any $i \sim j$ and $\int dx_k q_k(x_k) = 1$ for any $k$. An approximation of the form (6) together with the constraints on $q_{ij}$s and $q_k$s is called a Bethe approximation. Let us denote the family of such functions by $\mathcal{F}_B$. By choosing $q_{ij}(x_i, x_j) = q_i(x_i)q_j(x_j)$ one can easily check that $\mathcal{F}_{\mathrm{MF}} \subset \mathcal{F}_B$, thus $\mathcal{F}_B$ is non-empty. Assuming that the approximate marginals are correct and $q$ normalizes to 1 and then substituting (6) into (5), we get an approximation of the Kullback–Leibler divergence in (5) called the Bethe free energy.

Due to the factorization of $p$, we can write the Bethe free energy as

$$\begin{aligned} F_B(\{q_{ij}, q_k\}) = &- \sum_{i\sim j} \int d\boldsymbol{x}_{i,j}\, q_{ij}(\boldsymbol{x}_{i,j}) \log \Psi_{ij}(\boldsymbol{x}_{i,j}) \\ &+ \sum_{i\sim j} \int d\boldsymbol{x}_{i,j}\, q_{ij}(\boldsymbol{x}_{i,j}) \log \left[\frac{q_{ij}(\boldsymbol{x}_{i,j})}{q_i(x_i)q_j(x_j)}\right] + \sum_k \int dx_k\, q_k(x_k) \log q_k(x_k). \end{aligned} \tag{7}$$

One can also define the free energy through the Bethe approximation

$$\begin{aligned} \int d\boldsymbol{x}\, q\left(\boldsymbol{x}\right) \log q\left(\boldsymbol{x}\right) \approx &\sum_{i\sim j} \int d\boldsymbol{x}_{i,j}\, q\left(\boldsymbol{x}_{i,j}\right) \log q\left(\boldsymbol{x}_{i,j}\right) \\ &+ \sum_k \left(1 - n_k\right) \int dx_k\, q\left(x_k\right) \log q\left(x_k\right) \end{aligned}$$

of the entropy (e.g., Yedidia et al., 2000) and substitute the marginals with functions $q_{ij}$ and $q_k$ that normalize to one and are connected through the marginal consistency constraints $\int dx_j q_{ij}(x_i, x_j) = q_i(x_i)$.

From the stationary conditions of the Lagrangian corresponding to the fractional Bethe free energy (7) and the marginal consistency and normalization constraints, one can derive the same iterative algorithm as in (3) for the corresponding Lagrange multipliers of the consistency constraints (Yedidia et al., 2000). Similarly, approximate marginals can then be computed according to (4). It can be shown that there is a one-to-one correspondence between the stationary points of the Bethe free energy (7) and the fixed points of the message passing algorithm (3). Later, in Section 4 we will link the stable fixed points of (3) to the local minima of (7).

### 2.2 Fractional Free Energies and the Message Passing Algorithm

As mentioned in the introduction, in case of Gaussian models the message passing algorithm does not always converge. The reason for this appears to be that the approximate marginals may get indefinite or negative definite covariance matrices. Welling and Teh (2001) pointed out that this can be due to the unboundedness of the Bethe free energy.

Since $F_{\mathrm{M}F}$ is convex and bounded and the Bethe free energy might be unbounded, it seems plausible to analyze the fractional Bethe free energy

$$F_{\boldsymbol{\alpha}}(\{q_{ij}, q_k\}) = -\sum_{i\sim j} \int d\boldsymbol{x}_{i,j}\, q_{ij}(\boldsymbol{x}_{i,j}) \log \Psi_{ij}(\boldsymbol{x}_{i,j}) \tag{8}$$

$$+ \sum_{i\sim j} \frac{1}{\alpha_{ij}} \int d\boldsymbol{x}_{i,j}\, q_{ij}(\boldsymbol{x}_{i,j}) \log \left[\frac{q_{ij}(\boldsymbol{x}_{i,j})}{q_i(x_i) q_j(x_j)}\right] + \sum_k \int dx_k\, q_k(x_k) \log q_k(x_k).$$

introduced by Wiegerinck and Heskes (2003). Here, $\boldsymbol{\alpha}$ denotes the set of positive reals $\{\alpha_{ij}\}$. They showed that the fractional Bethe free energy "interpolates" between the mean field and the Bethe approximation. That is, for $\alpha_{ij} = 1$ we get the Bethe free energy, while in the case when all $\alpha_{ij}$s tend to 0, the mutual information between variables $x_i$ and $x_j$ is highly penalized, therefore, (8) enforces solutions close to the mean field solution. They also showed that the fractional message passing algorithm derived from (8) can be interpreted as Pearl's message passing algorithm with the difference that instead of computing local marginals—like in Pearl's algorithm—one computes local $\alpha_{ij}$–marginals.[3] The local $\alpha_{ij}$–marginals correspond to "true" local marginals when $\alpha_{ij} = 1$ and to local mean field approximations when $\alpha_{ij} = 0$. The resulting algorithm is called the fractional message passing algorithm and the message updates are defined as

$$\mu^{new}_{i\leftarrow j}(x_i)^{\alpha} = \int dx_j\, \Psi^{\alpha}_{ij}(x_i, x_j) \prod_{k\in\partial j\setminus i} \mu_{j\leftarrow k}(x_j)\, \mu_{j\leftarrow i}(x_j)^{1-\alpha}, \tag{9}$$

while the approximate marginals are computed according to

$$q_{ij}(x_i, x_j) \propto \Psi^{\alpha}_{ij}(x_i, x_j) \prod_{l\in\partial i\setminus j} \mu_{i\leftarrow l}(x_i)\, \mu_{i\leftarrow j}(x_i)^{1-\alpha} \prod_{k\in\partial j\setminus i} \mu_{j\leftarrow k}(x_j)\, \mu_{j\leftarrow i}(x_j)^{1-\alpha}. \tag{10}$$

3. We define the $\alpha$–marginals of a distribution $p$ as $\mathrm{argmin}_{\{q_k\}} D_{\alpha}\left[p \,\|\, \prod_k q_k\right]$, where $D_{\alpha}$ is the $\alpha$–divergence $\mathrm{D}_{\alpha}[p\,||\,q] = \left[\int d\boldsymbol{x} p(\boldsymbol{x})^{\alpha} q(\boldsymbol{x})^{1-\alpha} + \alpha \int d\boldsymbol{x} p(\boldsymbol{x}) + (1-\alpha)\int d\boldsymbol{x} q(\boldsymbol{x})\right]/\alpha(1-\alpha)$ (e.g., Minka, 2005).

Power expectation propagation by Minka (2004) is an approximate inference method that uses local approximations with $\alpha$–divergences. In case of Gaussian models power expectation propagation—with a fully factorized approximating distribution—leads to the same message passing algorithm as the one derived from (8) and the appropriate constraints. Starting from the idea of creating an upper bound on the log partition function when $p$ and $q$ are exponential distributions, Wainwright et al. (2003) derived a form of (8) where the $\alpha_{ij}$s are chosen such that this bound is convex in $\{q_{ij}, q_k\}$.

Message passing works well in practice, however, there are other ways to find the local minima of the fractional free energies like the direct minimization w.r.t. some parameterization of the approximate marginals $q_{ij}$ and $q_k$ (Welling & Teh, 2001). The latter method is slower but more likely to converge. In the following we analyze the Bethe free energy when expressed in terms of the moment parameters of the approximate marginals $q_{ij}$. Later in Section 4 we analyze the stability conditions of the fractional message passing algorithm and by expressing these conditions in term of the moment parameters of the approximate marginals, we show that stable fixed points of the fractional Gaussian message passing are local minima of the fractional Bethe free energy.

## 3. Bounds on the Gaussian Bethe Free Energy

In this section we analyze the parametric form of (8). We show that the fractional Gaussian Bethe free energy is a non-increasing function of $\boldsymbol{\alpha}$. By letting all $\alpha_{ij}$ tend to infinity, we obtain a lower bound for the free energies. It turns out that the condition for the lower bound to be bounded from below is the same as the pairwise normalizability condition in the work of Malioutov et al. (2006).

As mentioned in Section 2, without loss of generality, we can work with a unit diagonal $\boldsymbol{Q}$. We define $\boldsymbol{R}$ to be a matrix with zeros on its diagonal and $\boldsymbol{Q} = \boldsymbol{I} + \boldsymbol{R}$, where $\boldsymbol{I}$ is the identity matrix. $|\boldsymbol{R}|$ will be the matrix formed by the absolute values of $\boldsymbol{R}$'s elements. We use the moment parameterization $q_{ij}(\boldsymbol{x}_{i,j}) = N(\boldsymbol{x}_{i,j}|\boldsymbol{m}_{ij}, \boldsymbol{V}_{ij})$ and $q_k(x_k) = N(x_k|m_k, v_k)$, where $\boldsymbol{m}_{ij} = (m^i_{ij}, m^j_{ij})^T$ and $\boldsymbol{V}_{ij} = [v^i_{ij}, v_{ij}; v_{ji}, v^j_{ij}]$, with $v_{ij} = v_{ji}$. By using $m_i \equiv m^i_{ij} = m^i_{ik}$ and $v_i \equiv v^i_{ij} = v^k_{ik}$ for all $i \sim j$ and $i \sim k$, we embed the marginalization ($\int dx_j q_{ij}(x_i, x_j) = q_i(x_i)$ for all $i \sim j$) and normalization ($\int dx_j q_j(x_j) = 1$) constraints into the parameterization. With a slight abuse of notation the matrix formed by diagonal elements $v_k$ and off-diagonal elements $v_{ij}$ is denoted by $\boldsymbol{V}$ (we can take $v_{ij} = 0$ for all $i \nsim j$), the vector of means by $\boldsymbol{m} = (m_1, \ldots, m_n)^T$ and the vector of variances by $\boldsymbol{v} = (v_1, \ldots, v_n)^T$. Substituting $q_{ij}$ and $q_k$ into (8) one gets

$$
\begin{aligned}
F_{\boldsymbol{\alpha}}(\boldsymbol{m}, \boldsymbol{V}) = & -\boldsymbol{h}^T\boldsymbol{m} + \frac{1}{2}\boldsymbol{m}^T\boldsymbol{Q}\boldsymbol{m} + \frac{1}{2}\mathrm{tr}(\boldsymbol{Q}^T\boldsymbol{V}) \\
& - \frac{1}{2}\sum_{i\sim j}\frac{1}{\alpha_{ij}}\log\left(1 - \frac{v_{ij}^2}{v_i v_j}\right) - \frac{1}{2}\sum_k \log(v_k) + C,
\end{aligned} \tag{11}
$$

where $C$ is an irrelevant constant. Note that the variables $\boldsymbol{m}$ and $\boldsymbol{V}$ are independent, hence the minimizations of $F_{\boldsymbol{\alpha}}(\boldsymbol{m}, \boldsymbol{V})$ with regard to $\boldsymbol{m}$ and $\boldsymbol{V}$ can be carried out independently.

**Property 1.** *$F_{\boldsymbol{\alpha}}(\boldsymbol{m}, \boldsymbol{V})$ is convex and bounded in $(\boldsymbol{m}, \{v_{ij}\}_{i\neq j})$ and at any stationary point we have*

$$\begin{aligned} \boldsymbol{m}^* &= \boldsymbol{Q}^{-1}\boldsymbol{h} \\ {v_{ij}}^* &= -\mathrm{sign}(R_{ij})\frac{\sqrt{1+(2\alpha_{ij}R_{ij})^2 v_i v_j}-1}{2\alpha_{ij}|R_{ij}|}. \end{aligned} \tag{12}$$

*Proof:* $\boldsymbol{Q}$ is positive definite by definition, therefore, the quadratic term in $\boldsymbol{m}$ is convex and bounded. The variables $\boldsymbol{m}$ and $\boldsymbol{V}$ are independent and the minimum with regard to $\boldsymbol{m}$ is achieved at $\boldsymbol{m}^* = \boldsymbol{Q}^{-1}\boldsymbol{h}$. One can check that the second order derivative of $F_{\boldsymbol{\alpha}}(\boldsymbol{m}, \boldsymbol{V})$ with regard to $v_{ij}$ is non-negative and the first order derivative has only one solution when $-v_i v_j \leq v_{ij}^2 \leq v_i v_j$. Since the variables $v_{ij}$ are independent, one can conclude that $F_{\boldsymbol{\alpha}}(\boldsymbol{m}, \boldsymbol{V})$ is convex in $v_{ij}$. From the independence of $\boldsymbol{m}$ and $\boldsymbol{V}$, it follows that $F_{\boldsymbol{\alpha}}$ is convex in $(\boldsymbol{m}, \{v_{ij}\}_{i\neq j})$. □

Since the $\boldsymbol{V}_{ij}$s are constrained to be covariance matrices, we have $v_i v_j > v_{ij}^2$, thus the first logarithmic term in (11) is negative. As a consequence,

$$F_{\boldsymbol{\alpha}_1}(\boldsymbol{m}, \boldsymbol{V}) \geq F_{\boldsymbol{\alpha}_2}(\boldsymbol{m}, \boldsymbol{V}) \quad \text{for any} \quad \boldsymbol{0} < \boldsymbol{\alpha}_1 \leq \boldsymbol{\alpha}_2,$$

where $\boldsymbol{\alpha}_1 \leq \boldsymbol{\alpha}_2$ is taken element by element. This observation leads to the following property.

**Property 2.** *With $\alpha_{ij} = \alpha$, $F_\alpha$ is a non-increasing function of $\alpha$.*

Using Property 1 and substituting $v_{ij}^*$ into $F_{\boldsymbol{\alpha}}$ we define the constrained function

$$\begin{aligned} F^c_{\boldsymbol{\alpha}}(\boldsymbol{m}, \boldsymbol{v}) = -\boldsymbol{h}^T\boldsymbol{m} + \frac{1}{2}\boldsymbol{m}^T\boldsymbol{Q}\boldsymbol{m} + \frac{1}{2}\sum_k v_k \\ -\frac{1}{2}\sum_{i\sim j}\frac{1}{\alpha_{ij}}\left(\sqrt{1+(2\alpha_{ij}R_{ij})^2 v_i v_j}-1\right) \\ -\frac{1}{2}\sum_{n(i,j)}\frac{1}{\alpha_{ij}}\log\left(2\frac{\sqrt{1+(2\alpha_{ij}R_{ij})^2 v_i v_j}-1}{(2\alpha_{ij}R_{ij})^2 v_i v_j}\right) \\ -\frac{1}{2}\sum_k \log(v_k) + C^c, \end{aligned} \tag{13}$$

where $C^c$ is an irrelevant constant. From Property 2, it follows that when choosing $\alpha_{ij} = \alpha$, the function in (13) is a non-increasing function of $\alpha$. It then makes sense to take $\alpha \to \infty$ and verify whether we can get a lower bound for (13).

**Lemma 1.** *For any $\boldsymbol{v} > 0$, $0 \leq \alpha_1 \leq 1$ and $\alpha_2 \geq 1$ the following inequalities hold.*

$$\begin{aligned} F_{\mathrm{MF}}(\boldsymbol{m}, \boldsymbol{v}) &\geq F^c_{\alpha_1}(\boldsymbol{m}, \boldsymbol{v}) \geq F_B\left(\boldsymbol{m}, \{v_{ij}^*\}, \boldsymbol{v}\right) \\ F_B\left(\boldsymbol{m}, \{v_{ij}^*\}, \boldsymbol{v}\right) &\geq F^c_{\alpha_2}(\boldsymbol{m}, \boldsymbol{v}) \ldots \\ \ldots &\geq F_{\mathrm{MF}}(\boldsymbol{m}, \boldsymbol{v}) - \frac{1}{2}\sqrt{\boldsymbol{v}}^T|\boldsymbol{R}|\sqrt{\boldsymbol{v}} \end{aligned}$$

*Moreover, they are tight, that is,*

$$\lim_{\alpha\to 0} F_\alpha\left(\boldsymbol{m}, \{v_{ij}^*(\alpha)\}, \boldsymbol{v}\right) = F_{\mathrm{MF}}(\boldsymbol{m}, \boldsymbol{v})$$

*and*

$$\lim_{\alpha\to\infty} F_\alpha\left(\boldsymbol{m}, \{v^*_{ij}(\alpha)\}, \boldsymbol{v}\right) = F_{\mathrm{MF}}\left(\boldsymbol{m}, \boldsymbol{v}\right) - \frac{1}{2}\sqrt{\boldsymbol{v}}^T |\boldsymbol{R}| \sqrt{\boldsymbol{v}}.$$

*Proof:* Since the Bethe free energy is the specific case of the fractional Bethe free energy for $\alpha = 1$, the inequalities on $F_B(\boldsymbol{m}, \{v^*_{ij}(\alpha)\}, \boldsymbol{v})$ follow from Property 2. Now, we show that the upper and lower bounds are tight. The function $(1 + x^2)^{1/2} - 1$ behaves as $\frac{1}{2}x^2$ in the neighborhood of 0, therefore,

$$\lim_{\alpha\to 0} v^*_{ij}(\alpha) = 0 \qquad \text{and} \qquad \lim_{\alpha\to 0} \frac{\log\left(1 - \frac{v^{*\,2}_{ij}(\alpha)}{v_i v_j}\right)}{\alpha} = -\frac{1}{v_i v_j} \lim_{\alpha\to 0} \frac{v^{*\,2}_{ij}(\alpha)}{\alpha} = 0,$$

showing that $F_{\mathrm{MF}}(\boldsymbol{m}, \boldsymbol{v})$ is a tight upper bound.
As $\alpha$ tends to infinity, we have

$$\lim_{\alpha\to\infty} \frac{\sqrt{1 + (2\alpha R_{ij})^2 v_i v_j} - 1}{2\alpha} = |R_{ij}| \sqrt{v_i}\sqrt{v_j}$$

and

$$\lim_{\alpha\to\infty} \frac{1}{\alpha} \log\left(\frac{\sqrt{1 + (2\alpha R_{ij})^2 v_i v_j} - 1}{(2\alpha R_{ij})^2 v_i v_j}\right) = 0,$$

yielding a tight lower bound

$$\lim_{\alpha\to\infty} F_\alpha\left(\boldsymbol{m}, \{v^*_{ij}(\alpha)\}, \boldsymbol{v}\right) = F_{\mathrm{MF}}\left(\boldsymbol{m}, \boldsymbol{v}\right) - \frac{1}{2}\sqrt{\boldsymbol{v}}^T |\boldsymbol{R}| \sqrt{\boldsymbol{v}}. \qquad \square$$

Let $\lambda_{max}(|\boldsymbol{R}|)$ be the largest eigenvalue of $|\boldsymbol{R}|$. Analyzing the boundedness of the lower bound, we arrive at the following theorem.

**Theorem 1.** *For the fractional Bethe free energy in (11) corresponding to a connected Gaussian model, the following statements hold*

*(1) if* $\lambda_{max}(|\boldsymbol{R}|) < 1$*, then* $F_{\boldsymbol{\alpha}}$ *is bounded from below for all* $\boldsymbol{\alpha} > \boldsymbol{0}$*,*

*(2) if* $\lambda_{max}(|\boldsymbol{R}|) > 1$*, then* $F_{\boldsymbol{\alpha}}$ *is unbounded from below for all* $\boldsymbol{\alpha} > \boldsymbol{0}$*,*

*(3) if* $\lambda_{max}(|\boldsymbol{R}|) = 1$*, then* $F_{\boldsymbol{\alpha}}$ *is bounded from below if and only if* $\sum_i \sum_{i\sim j} \alpha_{ij}^{-1} \geq 2n$*.*

*Proof:* Since in $F_{\boldsymbol{\alpha}}$ there is no interaction between the parameters $\boldsymbol{m}$ and $\boldsymbol{V}$ and the term depending on $\boldsymbol{m}$ is bounded from below due to the positive definiteness of $\boldsymbol{Q}$, we can simply neglect this term when analyzing the boundedness of $F_{\boldsymbol{\alpha}}$. Let us write out in detail the lower bound of the fractional Bethe free energies in the form

$$F_{\mathrm{MF}}\left(\boldsymbol{m}, \boldsymbol{v}\right) - \frac{1}{2}\sqrt{\boldsymbol{v}}^T |\boldsymbol{R}| \sqrt{\boldsymbol{v}} = \tag{14}$$
$$\frac{1}{2}\boldsymbol{m}^T \boldsymbol{Q}^{-1} \boldsymbol{m} - \boldsymbol{h}^T \boldsymbol{m} + \frac{1}{2}\sqrt{\boldsymbol{v}}^T \left(\boldsymbol{I} - |\boldsymbol{R}|\right) \sqrt{\boldsymbol{v}} - \frac{1}{2}\boldsymbol{1}^T \log(\boldsymbol{v}) + \text{const.}$$

*Statement (1):* The condition $\lambda_{max}(|\boldsymbol{R}|) < 1$ implies that $\boldsymbol{I} - |\boldsymbol{R}|$ is positive definite. Now, $\log(x) \leq x - 1$, thus $\frac{1}{2}\sqrt{\boldsymbol{v}}^T(\boldsymbol{I} - |\boldsymbol{R}|)\sqrt{\boldsymbol{v}} - \boldsymbol{1}^T \log(\sqrt{\boldsymbol{v}}) \geq \frac{1}{2}\sqrt{\boldsymbol{v}}^T(\boldsymbol{I} - |\boldsymbol{R}|)\sqrt{\boldsymbol{v}} - \boldsymbol{1}^T\sqrt{\boldsymbol{v}} + n$. The latter is bounded from below and so it follows that (14) is bounded from below as well. According to Lemma 1, the boundedness of (14) implies that all fractional Bethe free energies are bounded from below.
*Statement (2):* We assumed that the Gaussian network is connected and undirected. According to the Perron-Frobenius theory of non-negative matrices (e.g., Horn & Johnson, 2005), $|\boldsymbol{R}|$ has a simple maximal eigenvalue $\lambda_{max}(|\boldsymbol{R}|)$ and all elements of the eigenvector $\boldsymbol{u}_{max}$ corresponding to it are positive. Let us take the fractional Bethe free energy and analyze its behavior when $\sqrt{\boldsymbol{v}} = t\boldsymbol{u}_{max}$ and $t \to \infty$. For large values of $t$ we have $(1 + (2\alpha_{ij}R_{ij})^2(u^i_{max}u^j_{max})^2t^4)^{1/2} \simeq 2\alpha_{ij}|R_{ij}|u^i_{max}u^j_{max}t^2$, therefore, the sum of the second and third term in (13) simplifies to $(1 - \lambda_{max}(|\boldsymbol{R}|))t^2$ and this term dominates over the logarithmic ones as $t \to \infty$. As a result, the limit is independent of the choice of $\alpha_{ij}$ and it tends to $-\infty$ whenever $\lambda_{max}(|\boldsymbol{R}|) > 1$.
*Statement (3):* If $\lambda_{max}(|\boldsymbol{R}|) = 1$, then the only direction in which the quadratic term will not dominate is $\sqrt{\boldsymbol{v}} = t\boldsymbol{u}_{max}$. Therefore, we have to analyze the behavior of the logarithmic terms in (13) when $t \to \infty$. For large $t$s these behave as $(\sum_{i\sim j}\alpha_{ij}^{-1} - 2n)\log(t)$. For this reason, the boundedness of $F^c_{\boldsymbol{\alpha}}$—and thus of $F_{\boldsymbol{\alpha}}$—depends on the condition in statement (3). □

It was shown by Malioutov et al. (2006) that the condition $\lambda_{max}(|\boldsymbol{R}|) < 1$ is an equivalent condition to pairwise normalizability. Therefore, pairwise normalizability is not only a sufficient condition for the message passing algorithm to converge, but it is also a necessary condition for the fractional Gaussian Bethe free energies to be bounded. Using Lemma 1, we can show that for a suitably chosen $\epsilon > 0$ there always exists an $\alpha_\epsilon$ such that the constrained fractional free energy $F^c_\alpha$ possesses a local minimum for any $0 < \alpha < \alpha_\epsilon$ (Property A2 in Section A of the Appendix).

**Example** In the case of models with an adjacency matrix (non-zero entries of $\boldsymbol{R}$) corresponding to a K–regular graph[4] and equal interaction weights $R_{ij} = r$, the maximal eigenvalue of $|\boldsymbol{R}|$ is $\lambda_{max}(|\boldsymbol{R}|) = Kr$ and the eigenvector corresponding to this eigenvalue is $\boldsymbol{1}$. (We define $\boldsymbol{1}$ as the vector that has all its elements equal to 1.) The model is symmetric and by verifying the stationary point conditions, it turns out that for some choice of $r$ and $\alpha$ there exists a local minimum, which also lies in the direction $\boldsymbol{1}$. One can show that when the model is not pairwise normalizable ($Kr > 1$), the critical $r$ below which the fractional Bethe free energy possesses this local minimum is $r_c(K,\alpha) = 1/2\sqrt{\alpha(K-\alpha)}$ and for any valid $r$ the critical $\alpha$ below which the fractional Bethe free energies possesses this local minimum is $\alpha_c(K,r) = \frac{1}{2}K(1 - \sqrt{1 - 1/(Kr)^2})$. These results are illustrated in Figure 2. (Note that for 2–regular graphs, all valid models are pairwise normalizable and possess a unique global minimum.) □

For K–regular graphs, the convexity of the fractional Bethe free energy in terms of $\{q_{ij}, q_k\}$ requires $\alpha \geq K$, a much stronger condition than $\alpha \geq \alpha_c(K,r)$. Thus, if we choose $\alpha$ sufficiently large such that the Bethe free energy is guaranteed to have a unique global minimum, this minimum is unbounded.

4. A K–regular graph is a graph in which all nodes are connected to K other nodes.

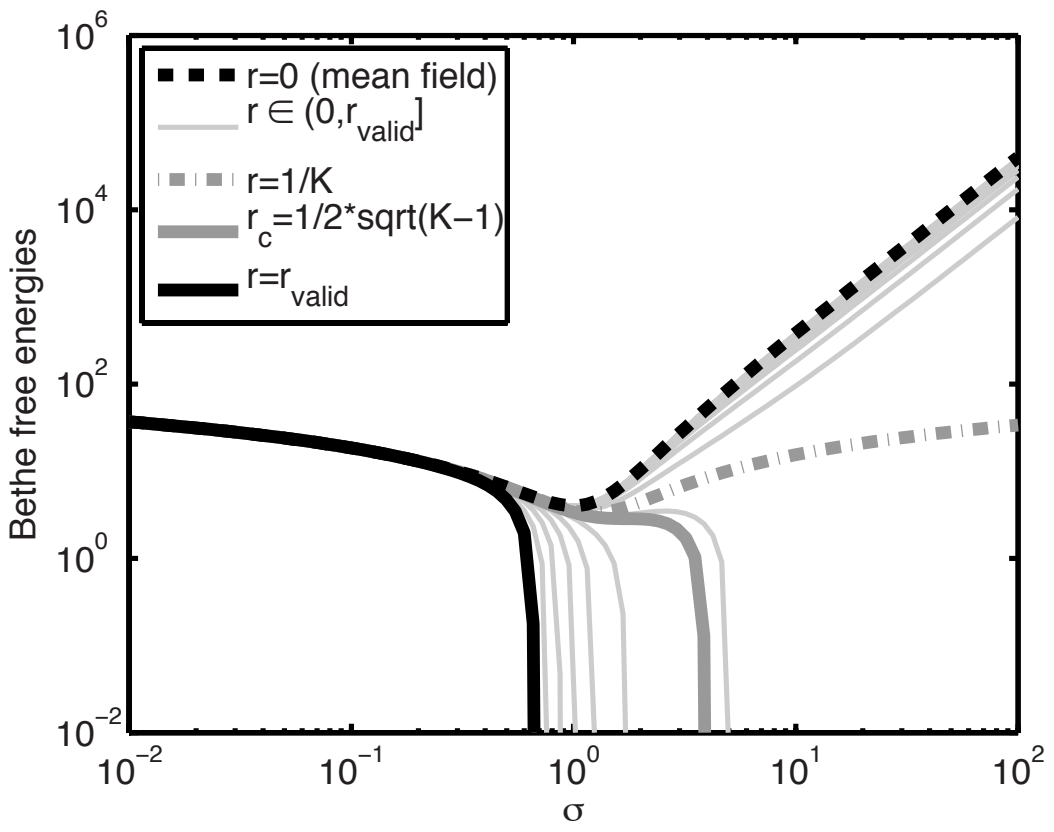

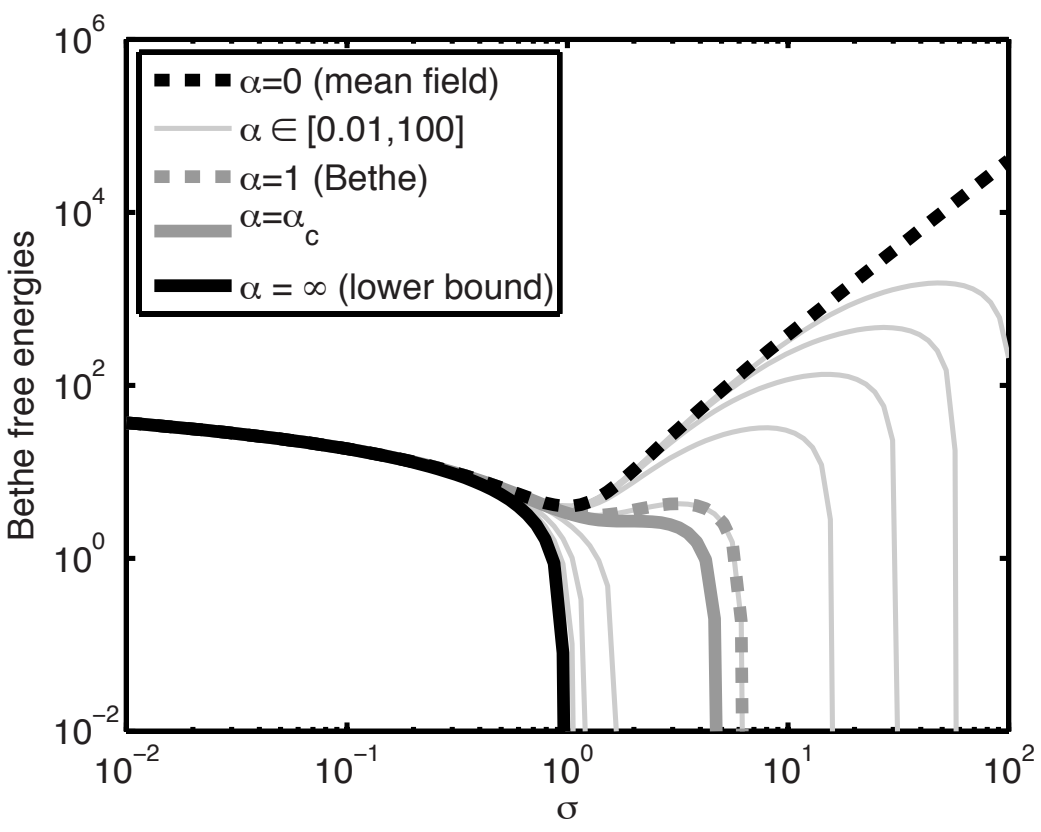

Figure 2: Visualizing critical parameters for a symmetric K-regular Gaussian model with $R_{ij} = r$. Plots in the left panel correspond to the constrained fractional Bethe free energies $F^c_\alpha$ for $\sqrt{\boldsymbol{v}} = \sigma\mathbf{1}$ for an 8 node 4–regular Gaussian model with r=0.27 ($Kr > 1$) and varying $\alpha$. Plots in the right panel correspond to the constrained Bethe free energies $F^c_1$ for $\sqrt{\boldsymbol{v}} = \sigma\mathbf{1}$ in an 8 node 4–regular Gaussian model with varying $r$. Here, $r_{\text{valid}}$ is the supremum of $r$s for which the model is valid, that is, $\boldsymbol{Q}$ is positive definite.

This example disproves the conjecture by Welling and Teh (2001), that is, even when the Bethe free energy is not bounded from below, it can possess a finite local minimum to which the message passing and the minimization algorithms can converge.

## 4. The Message Passing Algorithm in Gaussian Models

In this section, we turn our attention towards the properties of the message passing algorithm in Gaussian models. Following a similar line of argument as Watanabe and Fukumizu (2009) we show that stable fixed points of the message passing algorithm correspond to local minima of the Bethe free energy. We use the moment parameterization introduced in the previous sections. The way we proceed is the following: (1) we make a linear expansion of message passing iteration at a fixed point, (2) we express the linear expansion in terms of moment parameters corresponding to the fixed point and finally (3) we connect the properties of the latter with the properties of the Hessian of the Bethe free energy by using the matrix determinant lemma.

The form of the equation (9) implies that the messages $\mu_{i\leftarrow j}(x_i)$ are univariate Gaussian functions, thus we can express them in terms of two scalar (canonical) parameters $\eta_{ij}$ and $\lambda_{ij}$ such that $\log \mu_{i\leftarrow j}(x_i) = -\lambda_{ij} x_i^2/2 + \eta_{ij} x_i + \tau_{ijj}$, where the $\tau_{ij}$s are irrelevant constants. When expressed in terms of $\eta_{ij}$ and $\lambda_{ij}$, the damped message passing algorithm (9) translates

to

$$\eta_{ij}^{new} = (1-\epsilon)\eta_{ij} + \frac{\epsilon}{\alpha}\left[\alpha\gamma_{ij}^{i}h_i - \alpha R_{ij}\frac{\alpha\gamma_{ij}^{j}h_j + \sum\limits_{k\in\partial j\backslash i}\eta_{jk} + (1-\alpha)\eta_{ji}}{\alpha\gamma_{ij}^{j} + \sum\limits_{k\in\partial j\backslash i}\lambda_{jk} + (1-\alpha)\lambda_{ji}}\right] \tag{15}$$

$$\lambda_{ij}^{new} = (1-\epsilon)\lambda_{ij} + \frac{\epsilon}{\alpha}\left[\alpha\gamma_{ij}^{i} - \alpha^2 R_{ij}^2\left(\alpha\gamma_{ij}^{j} + \sum_{k\in\partial j\backslash i}\lambda_{jk} + (1-\alpha)\lambda_{ji}\right)^{-1}\right] \tag{16}$$

where $\gamma_{ij}^{i}, \gamma_{ij}^{j}, h_i$ and $R_{ij}$ are parameters of $\Psi_{ij}$ as in Section 2.1, with $R_{ij} = Q_{ij}$ and the assumption that $Q_{ii} = 1$. The approximate marginals $q_{ij}$ in (10) might not be normalizable, but the message passing iteration in (15) and (16) stays well defined unless there is a zero in the denominator on the rhs. This rarely happens in practice. However, it is more common that message passing converges while there are some intermediate steps at which the approximate marginals $q_{ij}$ are not normalizable. This can often be remedied by choosing an appropriate damping parameter $\epsilon$.

The iteration (16) for the $\lambda_{ij}$s is independent of $\eta_{ij}$s and the iteration (15) for the $\eta_{ij}$s is linear in $\eta_{ij}$. It is interesting to see that when $\boldsymbol{h} = \boldsymbol{0}$ neither the constrained Bethe free energy (13) nor the message passing algorithm (16) depend on the sign of $R_{ij}$. These are only relevant to compute the means—when $\boldsymbol{h} \neq \boldsymbol{0}$—and the signs of the correlations in (12). As a result, the marginal variances computed by either minimizing the Bethe free energy or by running the message passing algorithm can only depend on $|\boldsymbol{R}|$, similarly to the constrained fractional free energy $F_{\alpha}^{c}$.

### 4.1 Stability of the Gaussian Message Passing Algorithm

In the following we analyze the stability of the message passing iteration at its fixed points, that is, at the stationary points of the Lagrangian corresponding to the constrained minimization of the Gaussian Bethe free energy. We reiterate that we use $G = (V, E)$ to denote the graph corresponding to $\boldsymbol{Q}$, namely, $V = \{1, \ldots, n\}$ and $E = \{(i,j) : Q_{ij} \neq 0\}$. The vector $\boldsymbol{\lambda} \in \mathbb{R}^{|E|}$, corresponding to a set of messages $\{\lambda_{ij}\}_{ij}$, is composed by the concatenation of $\lambda_{ij}$s such that $ij$ is followed by $ji$ and the $(ij, ji)$ blocks follow a lexicographic order w.r.t. $ij$ and $i < j$. The vector $\boldsymbol{\eta}$ consists of the variables $\eta_{ij}$ and follows a similar structure as $\boldsymbol{\lambda}$. We define $\hat{\boldsymbol{r}}, \hat{\boldsymbol{h}}, \hat{\boldsymbol{\gamma}} \in \mathbb{R}^{|E|}$ as $\hat{r}_{ij} = \hat{r}_{ji} = R_{ij}$, $\hat{h}_{ij} = h_j$ and $\hat{\gamma}_{ij} = \gamma_{ij}^{j}$. We also define the $|E| \times |E|$ matrix

$$\mathcal{M}_{ij,kl}(\alpha) \equiv \begin{cases} 1 & \text{if } j = k \\ 1-\alpha & \text{if } kl = ji \\ 0 & \text{otherwise} \end{cases}$$

which encodes the weighted edge adjacency corresponding to $G$ and $\alpha$. The number of non-zero elements in $\boldsymbol{\mathcal{M}}(\alpha)$, scales roughly with $nnzeros(\boldsymbol{Q})^2/n$, where $nnzeros(\boldsymbol{Q})$ denotes the number of non-zeros in $\boldsymbol{Q}$. Since the parallel message update given Equations (15) and (16) can be rewritten in terms of two matrix-vector multiplications and element by element operations on vectors, the computational complexity of an update also scales as roughly with $nnzeros(\boldsymbol{Q})^2/n$.

With this notation, the local linearization of the update equations (15) and (16) can be written as

$$\frac{\partial\left(\boldsymbol{\eta}^{new}, \boldsymbol{\lambda}^{new}\right)}{\partial\left(\boldsymbol{\eta}, \boldsymbol{\lambda}\right)}(\boldsymbol{\eta}, \boldsymbol{\lambda}) = (1-\epsilon)\boldsymbol{I}\ldots$$
$$+\frac{\epsilon}{\alpha}\begin{bmatrix} -\mathrm{diag}\left(\alpha\hat{\boldsymbol{r}}\frac{1}{\alpha\hat{\boldsymbol{\gamma}}+\boldsymbol{\mathcal{M}}(\alpha)\boldsymbol{\lambda}}\right)\boldsymbol{\mathcal{M}}(\alpha) & \mathrm{diag}\left(\alpha\hat{\boldsymbol{r}}\frac{\alpha\hat{\boldsymbol{\gamma}}\hat{\boldsymbol{h}}+\boldsymbol{\mathcal{M}}(\alpha)\boldsymbol{\eta}}{(\alpha\hat{\boldsymbol{\gamma}}+\boldsymbol{\mathcal{M}}(\alpha)\boldsymbol{\lambda})^2}\right)\boldsymbol{\mathcal{M}}(\alpha) \\ \boldsymbol{0} & \mathrm{diag}\left(\alpha^2\hat{\boldsymbol{r}}^2\frac{1}{(\alpha\hat{\boldsymbol{\gamma}}+\boldsymbol{\mathcal{M}}(\alpha)\boldsymbol{\lambda})^2}\right)\boldsymbol{\mathcal{M}}(\alpha)\end{bmatrix}, \tag{17}$$

where all operations on vectors are element by element. The stability of a fixed point $(\boldsymbol{\eta}^*, \boldsymbol{\lambda}^*)$ depends on the union of the spectra of

$$\boldsymbol{J}_{\boldsymbol{\eta}}(\boldsymbol{\eta}^*, \boldsymbol{\lambda}^*) \equiv -\alpha^{-1}\mathrm{diag}\left(\alpha\hat{\boldsymbol{r}}(\alpha\hat{\boldsymbol{\gamma}} + \boldsymbol{\mathcal{M}}(\alpha)\boldsymbol{\lambda}^*)^{-1}\right)\boldsymbol{\mathcal{M}}(\alpha)$$

and

$$\boldsymbol{J}_{\boldsymbol{\lambda}}(\boldsymbol{\eta}^*, \boldsymbol{\lambda}^*) \equiv \alpha^{-1}\mathrm{diag}\left(\alpha^2\hat{\boldsymbol{r}}^2(\alpha\hat{\boldsymbol{\gamma}} + \boldsymbol{\mathcal{M}}(\alpha)\boldsymbol{\lambda}^*)^{-2}\right)\boldsymbol{\mathcal{M}}(\alpha).$$

It is important to point out that the stability properties depend only on $\boldsymbol{\lambda}^*$ and $\boldsymbol{R}$ and are independent of $\boldsymbol{\eta}^*$ and $\boldsymbol{h}$.

Our goal is to connect the stability properties of the message passing algorithm to the properties of the Bethe free energy. Therefore, we express the stability properties in terms of the moment parameters of approximate marginals. For any $\boldsymbol{\lambda}$ that leads to normalizable approximate marginals $q_{ij}(x_i, x_j)$, we can use (10) to identify the local covariance parameters $\boldsymbol{V}_{ij}$ defined in Section 3, but now without enforcing the marginal matching constraints $v_{ij}^i = v_{ik}^i$. The correspondence is given by

$$\begin{bmatrix} v_{ij}^i & v_{ij} \\ v_{ij} & v_{ij}^j \end{bmatrix}^{-1} = \frac{1}{v_{ij}^i v_{ij}^j - v_{ij}^2}\begin{bmatrix} v_{ij}^j & -v_{ij} \\ -v_{ij} & v_{ij}^i \end{bmatrix} \tag{18}$$
$$= \begin{bmatrix} \alpha\gamma_{ij}^i + \sum_{l\in\partial i\backslash j}\lambda_{il} + (1-\alpha)\lambda_{ij} & \alpha R_{ij} \\ \alpha R_{ij} & \alpha\gamma_{ij}^j + \sum_{k\in\partial j\backslash i}\lambda_{jk} + (1-\alpha)\lambda_{ji}\end{bmatrix}.$$

The approximate local covariances $v_{ij}$ are fully determined by $v_{ij}^i, v_{ij}^j$ and $r_{ij}$ and have the form as in (12). This leaves us with $|E|$ moment parameters to be computed by the message passing algorithm. Let $\hat{\boldsymbol{v}} \in \mathbb{R}^{|E|}$ be defined as $\hat{v}_{ij} = v_{ij}^i, \hat{v}_{ji} = v_{ij}^j$ and $y_{ij}(\hat{\boldsymbol{v}}) = v_{ij}/(v_{ij}^i v_{ij}^j - {v_{ij}}^2)$, where $v_{ij}$ is computed according to (12). It can be checked that the mapping between $\boldsymbol{y}$ and $\hat{\boldsymbol{v}}$ is continuous and bijective. This implies that the canonical to moment parameter transformation in (18) can be written as $\boldsymbol{y}(\hat{\boldsymbol{v}}) = \alpha\hat{\boldsymbol{\gamma}} + \boldsymbol{\mathcal{M}}(\alpha)\boldsymbol{\lambda}$. Since $\boldsymbol{\mathcal{M}}(\alpha)$ is singular only when $\alpha = K$ and the graph G is $K$-regular—see Property A1 in Section A of the Appendix for details—for the rest of the cases, there is a continuous, bijective mapping between the moment parameters $\hat{\boldsymbol{v}}$ and the canonical parameters $\boldsymbol{\lambda}$ that lead to normalizable approximate marginals.

At any fixed point $(\boldsymbol{\eta}^*, \boldsymbol{\lambda}^*)$ we have moment matching, that is, $v_{ij}^i = v_{ik}^i \equiv v_i^*$ for any $k, j \in \partial i$, therefore we can express the stability properties in terms of moment parameters $\boldsymbol{v}^* = (v_i^*, \ldots, v_n^*)$. Using (18) and defining the diagonal matrix $\boldsymbol{D} \in \mathbb{R}^{|E|\times|E|}$ with the diagonal elements $D_{ij,ij} = \sqrt{v_i^*}$, we get

$$\boldsymbol{D}\boldsymbol{J}_{\boldsymbol{\eta}}(\boldsymbol{\lambda}^*(\boldsymbol{v}^*))\boldsymbol{D}^{-1} = -\alpha^{-1}\mathrm{diag}\left(\frac{v_{ij}(\alpha, v_i^*, v_j^*)}{\sqrt{v_i^* v_j^*}}\right)\boldsymbol{\mathcal{M}}(\alpha) \tag{19}$$

and

$$\boldsymbol{D}^2\boldsymbol{J}_{\boldsymbol{\lambda}}(\boldsymbol{\lambda}^*(\boldsymbol{v}^*))\boldsymbol{D}^{-2} = \alpha^{-1}\mathrm{diag}\left(\frac{v_{ij}(\alpha, v_i^*, v_j^*)^2}{v_i^* v_j^*}\right)\boldsymbol{\mathcal{M}}(\alpha). \tag{20}$$

Let $\sigma(\boldsymbol{A})$ denote the spectrum of the matrix $\boldsymbol{A}$. Since we have $\sigma\left(\boldsymbol{D}\boldsymbol{J}_{\boldsymbol{\eta}}\boldsymbol{D}^{-1}\right) = \sigma\left(\boldsymbol{J}_{\boldsymbol{\eta}}\right)$ and $\sigma\left(\boldsymbol{D}^2\boldsymbol{J}_{\boldsymbol{\lambda}}\boldsymbol{D}^{-2}\right) = \sigma\left(\boldsymbol{J}_{\boldsymbol{\lambda}}\right)$, it is sufficient to analyze the spectral properties of the right hand sides in equations (19) and (20).

The message passing algorithm is asymptotically stable at $\boldsymbol{\lambda}^*(\boldsymbol{v}^*)$ if and only if

$$\max\left\{\rho\left(\boldsymbol{J}_{\boldsymbol{\eta}}(\boldsymbol{\lambda}^*(\boldsymbol{v}^*))\right), \rho\left(\boldsymbol{J}_{\boldsymbol{\lambda}}(\boldsymbol{\lambda}^*(\boldsymbol{v}^*))\right)\right\} < 1, \tag{21}$$

where $\rho(\cdot)$ denotes the spectral radius. It is interesting to see that although the functional forms of the free energies and the message passing algorithms are different in the Gaussian and discrete case, the stability conditions have similar forms. This will allow us to use some of the results of Watanabe and Fukumizu (2009). In the next section, we show the implications of this condition for the properties of the Hessian of the free energy.

### 4.2 Stable Fixed Points and Local Minima

The Hessian $\boldsymbol{H}[F_\alpha]$ of the Bethe free energy (11) depends only on the moment parameters $v_i, v_j$ and $v_{ij}$. Note that now, the $v_{ij}$s are unconstrained parameters. It is an $(|E|/2 + 2n) \times (|E|/2 + 2n)$ matrix and it has the form

$$\boldsymbol{H}[F_\alpha](\boldsymbol{V}) = \begin{bmatrix} \boldsymbol{Q} & \boldsymbol{0} & \boldsymbol{0} \\ \boldsymbol{0} & \mathrm{diag}\left(\frac{\partial^2 F_\alpha}{\partial^2 v_{ij}}\right) & \left[\frac{\partial^2 F_\alpha}{\partial v_{ij}\partial v_i}\right]_{ij,i} \\ \boldsymbol{0} & \left[\frac{\partial^2 F_\alpha}{\partial v_{ij}\partial v_i}\right]_{ij,i}^T & \left[\frac{\partial^2 F_\alpha}{\partial v_i \partial v_j}\right]_{i,j} \end{bmatrix},$$

where we use $\boldsymbol{V}$ to denote the collection of parameters $v_i$, $i = 1, \ldots, n$ and $v_{ij}$, $i \sim j$. Since the block corresponding to the partial differentials w.r.t. $v_{ij}$ is diagonal with positive elements, the Hessian is positive definite at $\boldsymbol{V}$ if the Schur complement corresponding to the partial differentials w.r.t. $v_i$s is positive definite at $\boldsymbol{V}$. The latter is given by

$$\begin{aligned} H^v_{ii}[F_\alpha](\boldsymbol{V}) &= \frac{\partial^2 F_\alpha}{\partial v_i \partial v_i} - \sum_{i\sim j} \left[\frac{\partial^2 F_\alpha}{\partial v_{ij}\partial v_i}\right]^2 \left[\frac{\partial F_\alpha}{\partial v_{ij}}\right]^{-1} \\ &= \frac{1}{2}\frac{1}{v_i^2}\left(1 + \frac{1}{\alpha}\sum_{i\sim j}\frac{c_{ij}^4}{1-c_{ij}^4}\right), \\ H^v_{ij}[F_\alpha](\boldsymbol{V}) &= \frac{\partial^2 F_\alpha}{\partial v_i \partial v_j} - \frac{\partial^2 F_\alpha}{\partial v_{ij}\partial v_i}\frac{\partial^2 F_\alpha}{\partial v_{ij}\partial v_j}\left[\frac{\partial^2 F_\alpha}{\partial^2 v_{ij}}\right]^{-1} \\ &= -\frac{1}{2}\frac{1}{v_i v_j}\frac{1}{\alpha}\frac{c_{ij}^2}{1-c_{ij}^4}, \end{aligned}$$

where we use the notation $c_{ij} = v_{ij}/\sqrt{v_i v_j}$.

Now, we would like to connect the condition in (21) to the positive definiteness of the matrix $\boldsymbol{H}^v[F_\alpha](\boldsymbol{V})$. In the following we show that stable fixed points $\boldsymbol{\lambda}^*(\boldsymbol{v}^*)$ of the Gaussian message passing algorithm, satisfying (21), correspond to local minima of the Gaussian free energy $F_\alpha$ at $\boldsymbol{v}^*$ and $v_{ij}(\alpha, v_i^*, v_j^*)$.

According to Watanabe and Fukumizu (2009), for any arbitrary vector $\boldsymbol{w} \in \mathbb{R}^{|E|}$ one has

$$\det\left(\boldsymbol{I}_{|E|} - \alpha^{-1}\text{diag}\left(\boldsymbol{w}\right)\boldsymbol{\mathcal{M}}(\alpha)\right) = \det\left(\boldsymbol{I}_n + \alpha^{-1}\boldsymbol{A}(\boldsymbol{w})\right)\prod_{ij}(1 - w_{ij}w_{ji}), \tag{22}$$

where

$$A_{ii}\left(\boldsymbol{w}\right) = \sum_{i\sim j}\frac{w_{ij}w_{ji}}{1-w_{ij}w_{ji}} \qquad \text{and} \qquad A_{ij}\left(\boldsymbol{w}\right) = -\frac{w_{ij}}{1-w_{ij}w_{ji}}. \tag{23}$$

The proof is an application of the matrix determinant lemma and a reproduction of it can be found in Section A of the Appendix. Equation (22) expresses the determinant of an $|E|\times|E|$ matrix as the determinant of an $n\times n$ matrix.

Let $\boldsymbol{c} \in \mathbb{R}^{|E|}$ with $c_{ij}(\boldsymbol{V}) = v_{ij}/\sqrt{v_i v_j}$. By substituting $\boldsymbol{w} = \boldsymbol{c}(\boldsymbol{V})^2$ in (23), we find that

$$\det\left(\boldsymbol{I} - \alpha^{-1}\text{diag}\left(\boldsymbol{c}(\boldsymbol{V})^2\right)\boldsymbol{\mathcal{M}}(\alpha)\right) = f\left(\boldsymbol{V}\right)\det\left(\boldsymbol{H}[F_\alpha](\boldsymbol{V})\right), \tag{24}$$

where $f\left(\boldsymbol{V}\right)$ is a positive function defined as

$$f\left(\boldsymbol{V}\right) = 2^n \alpha^{|E|}|\boldsymbol{Q}|^{-1}\prod_k v_k^2 \prod_{i\sim j}\frac{\left(v_i v_j - v_{ij}^2\right)^2}{v_i v_j + v_{ij}^2}\left(1 - \frac{v_{ij}^2}{v_i v_j}\right).$$

for all $\boldsymbol{V}$ corresponding to normalizable approximate marginals. Now, adapting the theorem of Watanabe and Fukumizu (2009) we have the following theorem.

**Theorem** *If* $\sigma\left(\alpha^{-1}\text{diag}\left(\boldsymbol{c}(\boldsymbol{V})^2\right)\boldsymbol{\mathcal{M}}(\alpha)\right) \subseteq \mathbb{C}\setminus\mathbb{R}_{\geq 1}$ *then the Hessian of the (Gaussian) Bethe free energy* $\boldsymbol{H}[F_\alpha]$ *is positive definite at* $\boldsymbol{V}$.

*Proof:* The assumption $\sigma\left(\alpha^{-1}\text{diag}\left(\boldsymbol{c}(\boldsymbol{V})^2\right)\boldsymbol{\mathcal{M}}(\alpha)\right) \subset \mathbb{C}\setminus\mathbb{R}_{\geq 1}$ implies that we have $\det\left(\boldsymbol{I} - \alpha^{-1}\text{diag}(\boldsymbol{c}(\boldsymbol{V})^2)\,\boldsymbol{\mathcal{M}}(\alpha)\right) > 0$. By choosing $V_{ij}(t) = tv_{ij}$ with $t \in [0,1]$, we find that $\boldsymbol{c}(\boldsymbol{V}(t))^2 = t^2\boldsymbol{c}(\boldsymbol{V})^2$, therefore, $\det\left(\boldsymbol{I} - \alpha^{-1}\text{diag}(\boldsymbol{c}(\boldsymbol{V}(t))^2)\boldsymbol{\mathcal{M}}(\alpha)\right) > 0$ for any $t \in [0,1]$.

This implies that $\det(\boldsymbol{H}[F_\alpha](\boldsymbol{V}(t))) > 0$ for any $t \in [0,1]$. Since $\boldsymbol{H}[F_\alpha](\boldsymbol{V}(0)) = \boldsymbol{I} > 0$ and the eigenvalues of $\boldsymbol{H}[F_\alpha](\boldsymbol{V}(t))$ change continuously w.r.t. $t \in [0,1]$, it results that $\boldsymbol{H}[F_\alpha](\boldsymbol{V}(1)) > 0$ for any $\boldsymbol{V}$, thus satisfying the condition of the theorem. □

The fixed point $(\boldsymbol{\eta}^*, \boldsymbol{\lambda}^*)$ is stable if and only if $\max\{\rho(\boldsymbol{J}_{\boldsymbol{\eta}}(\boldsymbol{\lambda}^*(\boldsymbol{v}^*))), \rho(\boldsymbol{J}_{\boldsymbol{\lambda}}(\boldsymbol{\lambda}^*(\boldsymbol{v}^*)))\} < 1$. This implies $\sigma\left(\alpha^{-1}\mathrm{diag}(\boldsymbol{c}(\boldsymbol{V}^*)^2)\boldsymbol{\mathcal{M}}(\alpha)\right) \subseteq \mathbb{C} \setminus \mathbb{R}_{\geq 1}$ and leads to the following property.

**Property 3.** *Stable fixed points $(\boldsymbol{\eta}^*, \boldsymbol{\lambda}^*)$ of the damped Gaussian message passing algorithm* (16) *are local minima of the Gaussian Bethe free energy $F^c_\alpha$ in* (13) *at $\boldsymbol{v}^*(\boldsymbol{\lambda}^*)$.*

The above shows that the boundedness of $F_\alpha$ or the existence of local minima in case of an unbounded $F_\alpha$ plays a significant role in the convergence of Gaussian message passing. We illustrate this in Section 5. If the fractional message passing algorithm converges then it converges to a set of messages that corresponds to a local minimum of the fractional free energy. This also implies that the mean parameters of the local approximate marginals are exact (see Property 1. in Section 3). Note that the observations in Section 3 and Property A2 in the Appendix together with Property 3 imply that there is always a range of $\alpha$ values for which the fractional free energy possesses a local minimum to which the fractional message passing can converge.

### 4.3 The Damping and the Fractional Parameters

The local stability condition in (21) is independent of the damping parameter $\epsilon$. Therefore, it does not alter the local stability properties, it only makes the iteration slower and numerically more stable, that is, it can dampen the possible periodic trajectories of the message passing algorithm.

The fractional parameter $\alpha$ characterizes the inference process and as we have seen in the example in the previous sections, by choosing smaller $\alpha$s we can create local minima. In the particular case when $\boldsymbol{h} = \boldsymbol{0}$, there is a somewhat similar property for the message passing updates as well. Let $\Lambda \in \mathbb{R}^{|E|}$ be the set of messages $\boldsymbol{\lambda}$ that lead to normalizable approximate marginals. The set $\Lambda$ is characterized by the model parameters $|\boldsymbol{R}|, \hat{\boldsymbol{\gamma}}$ and $\alpha$. We reiterate that the elements of $\hat{\boldsymbol{v}}$ are the local variances $v^i_{ij}$ and $v^j_{ij}$ and there is a continuous bijective mapping between $\boldsymbol{\lambda} \in \Lambda$ and $\hat{\boldsymbol{v}} \in \mathbb{R}^{|E|}_+$ given by $\boldsymbol{y}(\hat{\boldsymbol{v}}) = \alpha\hat{\boldsymbol{\gamma}} + \boldsymbol{\mathcal{M}}(\alpha)\boldsymbol{\lambda}$, unless $\alpha = K$ and $G$ is $K$-regular. This allows us to study the stability properties in terms of moment parameters $\hat{\boldsymbol{v}}(\boldsymbol{\lambda})$. Let $\boldsymbol{c}(\hat{\boldsymbol{v}}, \alpha) = [v_{ij}(\alpha, v^i_{ij}, v^j_{ij})/\sqrt{v^i_{ij}v^j_{ij}}]_{ij}$ be the vector of "local correlations". By using Gershgorin's theorem (Horn & Johnson, 2005) and $\boldsymbol{c}(\hat{\boldsymbol{v}}, \alpha)^2 \leq \boldsymbol{c}(\hat{\boldsymbol{v}}, \alpha)$, we find that for any eigenvalue $\beta$ of $\alpha^{-1}\mathrm{diag}(\boldsymbol{c}(\hat{\boldsymbol{v}}, \alpha))\boldsymbol{\mathcal{M}}(\alpha)$ or $\alpha^{-1}\mathrm{diag}(\boldsymbol{c}(\hat{\boldsymbol{v}}, \alpha))^2\boldsymbol{\mathcal{M}}(\alpha)$ we have

$$|\beta| \leq \max_{i,j}\left[\alpha^{-1}\boldsymbol{c}(\hat{\boldsymbol{v}}, \alpha)\left[(n_j - 1) + |1 - \alpha|\right]\right].$$

When $\boldsymbol{h} = \boldsymbol{0}$, there are no updates in $\boldsymbol{\eta}$, the rhs of the above equation depends on $\alpha^{-1}\boldsymbol{c}(\hat{\boldsymbol{v}}, \alpha)^2$ (see Equations (17) and (20)) and we have $\lim_{\alpha \to 0} \alpha^{-1}\boldsymbol{c}(\hat{\boldsymbol{v}}, \alpha)^2 = \boldsymbol{0}$, thus, small $\alpha$ values can help to achieve convergence. However, when $\boldsymbol{h} \neq \boldsymbol{0}$ the term $\alpha^{-1}\boldsymbol{c}(\hat{\boldsymbol{v}}, \alpha)$ is dominating and the effects of decreasing $\alpha$ towards zero can be ambiguous.

## 5. Experiments

We implemented both direct minimization and fractional message passing and analyzed their behavior for different values of $\lambda_{max}(|\boldsymbol{R}|)$. For reasons of simplicity, we set all $\alpha_{ij}$s equal. The results on an small scale model are summarized in Figure 3. Note that there is a good correspondence between the behavior of the fractional Bethe free energies in the direction of the eigenvalue corresponding to $\lambda_{max}(|\boldsymbol{R}|)$ and the convergence of the Newton method. The Newton method was started from different initial points. We experienced that when $\lambda_{max}(|\boldsymbol{R}|) > 1$ and setting the initial value to $\boldsymbol{v}_0 = t^2 \boldsymbol{u}_{max}^2$ , the algorithm did not converge for high values of $t$. This can be explained by the top plots in Figure 3: for high values of $t$, the initial point might not be in the convergence region of the local minimum. For the fractional message passing algorithm we used two types of initialization: (1) when $\lambda_{max}(|\boldsymbol{R}|) < 1$ we set $\Psi_{ij}$ such that they are all normalizable by setting $\gamma_{ij}^i = |R_{ij}| u_{max}^j / \lambda_{max} u_{max}^i$ (Malioutov et al., 2006), (2) when $\lambda_{max}(|\boldsymbol{R}|) \geq 1$, we used $\gamma_{ij}^i = 1/n_i$, that is, a symmetric partitioning of the diagonal elements. We set the initial messages such that all approximate marginals are normalizable in the first step of the iteration.

We experienced a behavior similar to that described by Welling and Teh (2001) for standard message passing, namely, fractional message passing and direct minimization either both converge or both fail to converge. Our experiments in combination with Theorem 1 show that when $\lambda_{max}(|\boldsymbol{R}|) > 1$, standard message passing at best converges to a local minimum of the Bethe free energy. If standard message passing fails to converge, one can decrease $\alpha$ and search for a stationary point—preferably a local minimum—of the corresponding fractional free energy.

It can be seen from the results in the right panels of Figure 2, that when the model is no longer pairwise normalizable, the local minimum and not the unbounded global minimum can be viewed the natural continuation of the (bounded) global minimum for pairwise normalizable models. This explains why the quality of the approximation at the local minimum for models that are not pairwise normalizable is still comparable to that at the global minimum for models that are pairwise normalizable.

## 6. Conclusions

As we have seen, $F_{\mathrm{M}F}$ and $F_{\mathrm{MF}} - \frac{1}{2}\sqrt{\boldsymbol{v}}^T |\boldsymbol{R}| \sqrt{\boldsymbol{v}}$ provide tight upper and lower bounds for the Gaussian fractional Bethe free energies. It turns out that pairwise normalizability is not only a sufficient condition for the message passing algorithm to converge, but it is also a necessary condition for the Gaussian fractional Bethe free energies to be bounded from below.

If the model is pairwise normalizable, then the lower bound is bounded, and both direct minimization and message passing are converging. In our experiments both converged to the same minimum. This suggests that in the pairwise normalizable case, fractional Bethe free energies possess a unique global minimum.

If the model is not pairwise normalizable, then none of the fractional Bethe free energies are bounded from below. However, there is always a range of $\alpha$ values for which the fractional free energy possesses a local minimum to which both direct minimization and fractional message passing can converge. Thus, by decreasing $\alpha$ towards zero, one gets closer to the mean field energy and a finite local minimum will appear (Property A2 in the Appendix). We experienced that for a suitable range of $\alpha$s,$\epsilon$s and initial values the fractional Gaussian message passing can be made to converge.

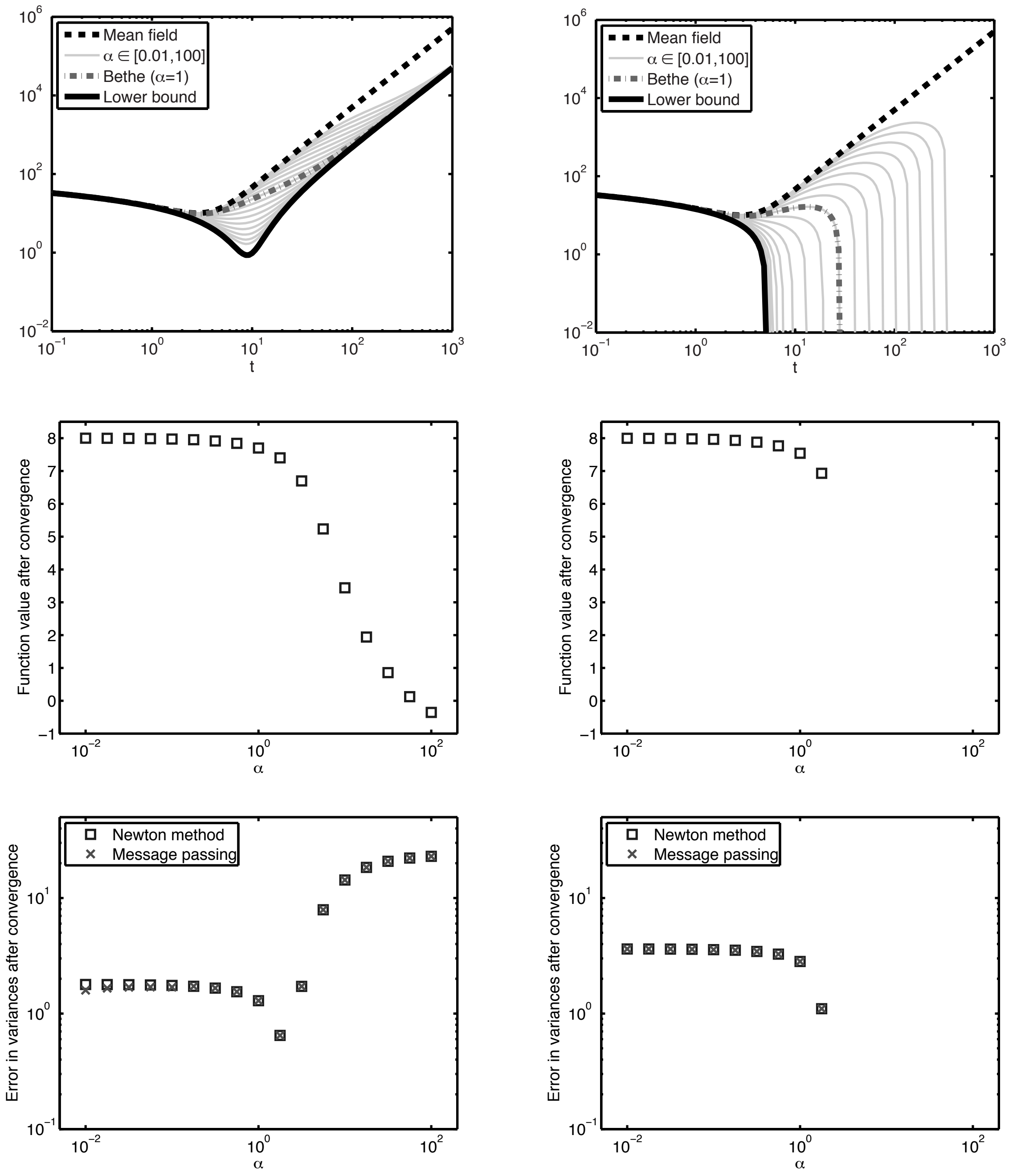

Figure 3: The top panels show the constrained fractional Bethe free energies of an Gaussian model with 8 variables in the direction $\sqrt{\boldsymbol{v}} = t\boldsymbol{u}_{max}$, where $\boldsymbol{u}_{max}$ is the eigenvector corresponding to $\lambda_{max}(|\boldsymbol{R}|)$ for $\lambda_{max}(|\boldsymbol{R}|) = 0.9$ (top-left) and $\lambda_{max}(|\boldsymbol{R}|) = 1.1$ (top-right). The thick lines are the functions $F_{\mathrm{MF}}$ (dashed), $F_B$ (dashed dotted) and the lower bound $F_{\mathrm{MF}} - \frac{1}{2}\sqrt{\boldsymbol{v}}^T|\boldsymbol{R}|\sqrt{\boldsymbol{v}}$ (continuous). The thin lines are the constrained $\alpha$-fractional free energies $F^c_\alpha$ for $\alpha \in [10^{-2}, 10^2]$. Center panels show the final function values after the convergence of the Newton method. The bottom panels show the $||\cdot||_2$ error in approximation for the single node standard deviations $\boldsymbol{\sigma} = \sqrt{\boldsymbol{v}}$. Missing values indicate non-convergence.

As mentioned in Section 2.1, $\alpha_{ij}$s correspond to using local $\alpha_{ij}$ divergences when applying power expectation propagation with a fully factorized approximating distribution. Seeger (2008) reports that when expectation propagation does not converge, applying power expectation propagation with $\alpha < 1$ helps to achieve convergence. In the case of the problem addressed in this paper this behavior can be explained by the observation that small $\alpha$s make a finite local minima more likely to occur and thus prevents the covariance matrices from becoming indefinite or even non positive definite. Although the most common reason for using $\alpha < 1$ in EP is numerical robustness, it also implies finding the saddle point of the $\alpha$-fractional EP free energy. It might be interesting to investigate whether it is the same reason why convergence is more likely as in the case of Gaussian fractional message passing.

Wainwright et al. (2003) propose to convexify the Bethe free energy for discrete models by choosing $\alpha_{ij}$s sufficiently large such that the fractional Bethe free energy has a unique global minimum. This strategy appears to fail for Gaussian models. Convexification makes the possibly useful finite local minima disappear, leaving just the unbounded global minimum. In the case of the more general hybrid models, the use of the convexification is still unclear.

The example in Section 3 disproves the conjecture in the work of Welling and Teh (2001): even when the Bethe free energy is not bounded from below, it can possess a finite local minimum to which the message passing and the minimization algorithms can converge.

We have shown that stable fixed points of the Gaussian fractional message passing algorithms are local minima of the fractional Bethe free energy. Although the existence of a local minimum does not guarantee the convergence of the message passing algorithm, in practice we experienced that the existence of a local minimum implies convergence. Based on these results, we *hypothesize* that when pairwise normalizability does not hold, the Gaussian Bethe free energy and the Gaussian message passing algorithm ($\alpha = 1$) can have two types of behavior:

(1) the Gaussian Bethe free energy possesses a unique finite local minimum to which optimization methods can converge by starting from, say, the mean field solution $v_i = 1/Q_{ii}$; the Gaussian message passing has a corresponding unique stable fixed point, to which it can converge with suitable starting point and sufficient damping,

(2) no finite local minimum exists, and thus, both the optimization and the message passing algorithm diverge.

By using the fractional free energy and the fractional message passing and by varying $\alpha$, one can switch between these behaviors. Computing the critical $\alpha_c(|\boldsymbol{R}|)$ for a general $|\boldsymbol{R}|$ remains an open question. We believe that the properties of the free energies in $K$-regular symmetric models (Section 3), where the critical values can be easily computed, give a good insight into the properties of the free energies for general Gaussian models.

## Acknowledgments

We would like to thank Jason K. Johnson for sharing his ideas about the properties of the message passing algorithm in $K$-regular models. We would also like to thank the anonymous reviewers for their valuable comments on earlier versions of the manuscript. The research reported in this paper was supported by VICI grant 639.023.604 from the Netherlands Organization for Scientific Research (NWO).

## Appendix A. Properties and Proofs

**Lemma A1.** (Watanabe & Fukumizu, 2009) *For any graph $G = (V, E)$, edge adjacency matrix $\boldsymbol{\mathcal{M}}(\alpha)$ (defined in Section 4.1), and arbitrary vector $\boldsymbol{w} \in \mathbb{R}^{|E|}$, one has*

$$\det\left(\boldsymbol{I}_{|E|} - \alpha^{-1}\mathrm{diag}\left(\boldsymbol{w}\right)\boldsymbol{\mathcal{M}}(\alpha)\right) = \det\left(\boldsymbol{I}_{|V|} + \alpha^{-1}\boldsymbol{A}(\boldsymbol{w})\right)\prod_{ij}(1 - w_{ij}w_{ji}),$$

*where*

$$A_{ii}\left(\boldsymbol{w}\right) = \sum_{i\sim j}\frac{w_{ij}w_{ji}}{1 - w_{ij}w_{ji}} \qquad \textit{and} \qquad A_{ij}\left(\boldsymbol{w}\right) = -\frac{w_{ij}}{1 - w_{ij}w_{ji}}.$$

*Proof*: We reproduce the proof in a somewhat simplified form. Let us define $\boldsymbol{U}_{ij,\cdot} = \boldsymbol{e}_j^T$, $\boldsymbol{V}_{ij,\cdot} = \boldsymbol{e}_i^T$—where $\boldsymbol{e}_k$ is the $k^{th}$ unit vector of $\mathbb{R}^n$—and $\boldsymbol{S}$ with

$$\begin{bmatrix} S_{ij,ij} & S_{ij,ji} \\ S_{ji,ij} & S_{ji,ji} \end{bmatrix} = \begin{bmatrix} 0 & 1 \\ 1 & 0 \end{bmatrix},$$

then we have $\boldsymbol{\mathcal{M}}(\alpha) = \boldsymbol{U}\boldsymbol{V}^T - \alpha\boldsymbol{S}$. Let us define $\boldsymbol{W} \in \mathbb{R}^{|E|\times|E|}$ a diagonal matrix with $w_{ij,ij} = w_{ij}$. Using the matrix determinant lemma this reads as

$$\begin{aligned}
&\det\left(\boldsymbol{I} - \alpha^{-1}\boldsymbol{W}\left(\boldsymbol{U}\boldsymbol{V}^T - \alpha\boldsymbol{S}\right)\right) \\
&\quad = \det\left(\boldsymbol{I} + \boldsymbol{W}\boldsymbol{S} - \alpha^{-1}\boldsymbol{W}\left(\boldsymbol{U}\boldsymbol{V}^T\right)\right) \\
&\quad = \det\Big(\boldsymbol{I} - \alpha^{-1}\boldsymbol{W}\left(\boldsymbol{U}\boldsymbol{V}^T\right)\left(\boldsymbol{I} + \boldsymbol{W}\boldsymbol{S}\right)^{-1}\Big)\det\left(\boldsymbol{I} + \boldsymbol{W}\boldsymbol{S}\right) \\
&\quad = \det\Big(\boldsymbol{I} - \alpha^{-1}\boldsymbol{V}^T\left(\boldsymbol{I} + \boldsymbol{W}\boldsymbol{S}\right)^{-1}\boldsymbol{W}\boldsymbol{U}\Big)\det\left(\boldsymbol{I} + \boldsymbol{W}\boldsymbol{S}\right).
\end{aligned}$$

The $(ij, ji)$ block of $\left(\boldsymbol{I} + \boldsymbol{W}\boldsymbol{S}\right)^{-1}\boldsymbol{W}$ is

$$\frac{1}{1 - w_{ji}w_{ji}}\begin{bmatrix} 1 & -w_{ij} \\ -w_{ji} & 1 \end{bmatrix}\begin{bmatrix} w_{ij} & 0 \\ 0 & w_{ji} \end{bmatrix} = \frac{1}{1 - w_{ji}w_{ji}}\begin{bmatrix} w_{ij} & -w_{ij}w_{ji} \\ -w_{ji}w_{ij} & w_{ji} \end{bmatrix}$$

and thus, we can define $\boldsymbol{A} \equiv \boldsymbol{V}^T\left(\boldsymbol{I} + \boldsymbol{W}\boldsymbol{S}\right)^{-1}\boldsymbol{W}\boldsymbol{U}$ such that

$$A_{i,i} = \sum_{i\sim j}\frac{w_{ij}w_{ji}}{1 - w_{ij}w_{ji}} \quad \text{and} \quad A_{i,j} = -\frac{w_{ij}}{1 - w_{ij}w_{ji}}.$$

This completes the proof of the matrix determinant lemma (22) in Section 4.2. □

**Property A1.** *The matrix* $\boldsymbol{\mathcal{M}}(\alpha) = \boldsymbol{U}\boldsymbol{V}^T - \alpha\boldsymbol{S}$ *is singular only for* $K$*-regular graphs with* $\alpha = K$.
*Proof:* Let $x \in \mathbb{R}^{|E|}$ and $\boldsymbol{y} = \boldsymbol{\mathcal{M}}(\alpha)\boldsymbol{x}$. Then $y_{ij} = \sum_{k\sim j} x_{jk} - \alpha x_{ji}$. Let us fix $j$, then $y_{ij} = 0$ for any $i$ means that $\sum_{k\sim j} x_{jk} = \alpha x_{ji}$ for any $i$. This can only hold if the graph is $K$-regular, $\alpha = K$ and all $x_{ij}$s are equal or $x_{ij} = 0$ for all pair indices $ij$. □

**Property A2.** *For a suitably chosen* $\epsilon > 0$*, there exists an* $\alpha_\epsilon$ *such that the constrained fractional free energy* $F^c_\alpha$ *possesses a local minimum for all* $0 < \alpha < \alpha_\epsilon$.
*Proof:* Let us define $\boldsymbol{v}^*_{MF} = \text{argmin}_{\boldsymbol{v}} F_{MF}(\boldsymbol{v})$ and

$$U^\epsilon_{MF} = \{\boldsymbol{v} : F_{MF}(\boldsymbol{v}) \leq F_{MF}(\boldsymbol{v}^*_{MF}) + 2\epsilon\}\,.$$

The form of $F_{MF}$ implies that we can always choose $\epsilon$ such that $U^\epsilon_{MF}$ is a proper subset of the positive "quadrant" in $\mathbb{R}^n$, in other words, $U^\epsilon_{MF} \subset \mathbb{R}^n_+$. Then due to the properties of $F_{MF}$ (continuous and convex, with a unique finite global minimum attained at a finite value), the domain $U^\epsilon_{MF}$ is closed, bounded, convex and $\boldsymbol{v}^*_{MF} \in U^\epsilon_{MF} \setminus \partial U^\epsilon_{MF}$, that is, $\boldsymbol{v}^*_{MF}$ is in the interior of $U^\epsilon_{MF}$. Since $F_{MF}$ and $F^c_\alpha(\boldsymbol{v})$ are continuous on $\mathbb{R}^n_+$, the set $U^\epsilon_{MF}$ is closed and bounded and $\lim\limits_{\alpha\to 0} F^c_\alpha(\boldsymbol{v}) = F_{MF}(\boldsymbol{v})$ (pointwise convergence) for all $\boldsymbol{v} \in \mathbb{R}^n_+$, it follows that $F^c_\alpha$ converges uniformly on $U^\epsilon_{MF}$ as $\alpha \to 0$. This, together with the monotonicity of $F^c_\alpha$ w.r.t. $\alpha$, implies that there exists $\alpha_\epsilon$ such that $F_{MF}(\boldsymbol{v}_{MF}) - \epsilon < F^c_\alpha(\boldsymbol{v}_{MF}) < F_{MF}(\boldsymbol{v}_{MF})$ for all $0 < \alpha < \alpha_\epsilon$ and all $\boldsymbol{v} \in U^\epsilon_{MF}$. Let us fix $\alpha$. It is known that, since $U^\epsilon_{MF}$ is closed and bounded and $F^c_\alpha$ is continuous, $F^c_\alpha$ attains its extrema on $U^\epsilon_{MF}$. Since $F_{MF}(\boldsymbol{v}) = F_{MF}(\boldsymbol{v}^*_{MF}) + 2\epsilon$ for all $\boldsymbol{v} \in \partial U^\epsilon_{MF}$ and $F^c_\alpha(\boldsymbol{v}) > F_{MF}(\boldsymbol{v}) - \epsilon$ for all $\boldsymbol{v} \in U^\epsilon_{MF}$ it follows that $F^c_\alpha(\boldsymbol{v}) > F_{MF}(\boldsymbol{v}^*_{MF}) + \epsilon$ for all $\boldsymbol{v} \in \partial U^\epsilon_{MF}$. We have chosen $\alpha$ such that $F_{MF}(\boldsymbol{v}^*_{MF}) - \epsilon < F^c_\alpha(\boldsymbol{v}^*_{MF}) < F_{MF}(\boldsymbol{v}^*_{MF})$. The latter two conditions imply that one of the extrema has to be a local minimum in the interior of $U^\epsilon_{MF}$. □

## References

Bickson, D. (2009). *Gaussian Belief Propagation: Theory and Application.* Ph.D. thesis, The Hebrew University of Jerusalem.

Cseke, B., & Heskes, T. (2008). Bounds on the Bethe free energy for Gaussian networks. In McAllester, D. A., & Myllymäki, P. (Eds.), *UAI 2008, Proceedings of the 24th Conference in Uncertainty in Artificial Intelligence*, pp. 97–104. AUAI Press.

Heskes, T. (2003). Stable fixed points of loopy belief propagation are minima of the Bethe free energy. In Becker, S., Thrun, S., & Obermayer, K. (Eds.), *Advances in Neural Information Processing Systems 15*, pp. 359–366, Cambridge, MA. The MIT Press.

Heskes, T., Opper, M., Wiegerinck, W., Winther, O., & Zoeter, O. (2005). Approximate inference techniques with expectation constraints. *Journal of Statistical Mechanics: Theory and Experiment*, *2005*, P11015.

Heskes, T. (2004). On the uniqueness of loopy belief propagation fixed points. *Neural Computation*, *16*, 2379–2413.

Horn, R. A., & Johnson, C. (2005). *Matrix Analysis.* Cambridge University Press, Cambridge, UK.

Jaakkola, T. (2000). Tutorial on variational approximation methods. In Opper, M., & Saad, D. (Eds.), *Advanced mean field methods: theory and practice*, pp. 129–160, Cambridge, MA. The MIT Press.

Johnson, J. K., Bickson, D., & Dolev, D. (2009). Fixing convergence of Gaussian belief propagation. *CoRR*, *abs/0901.4192*.

Malioutov, D., Johnson, J., & Willsky, A. (2006). Walk-sums and belief propagation in Gaussian graphical models. *Journal of Machine Learning Research*, *7*, 2031–2064.

Minka, T. P. (2004). Power EP. Tech. rep., Microsoft Research Ltd., Cambridge, UK, MSR-TR-2004-149.

Minka, T. P. (2005). Divergence measures and message passing. Tech. rep. MSR-TR-2005-173, Microsoft Research Ltd., Cambridge, UK.

Moallemi, C., & Roy, B. V. (2006). Consensus propagation. In Weiss, Y., Schölkopf, B., & Platt, J. (Eds.), *Advances in Neural Information Processing Systems 18*, pp. 899–906. MIT Press, Cambridge, MA.

Murphy, K., Weiss, Y., & Jordan, M. I. (1999). Loopy belief propagation for approximate inference: An empirical study. In *Proceedings of the Fifteenth Conference on Uncertainty in Artificial Intelligence*, Vol. 9, pp. 467–475, San Francisco, USA. Morgan Kaufman.

Nishiyama, Y., & Watanabe, S. (2009). Accuracy of loopy belief propagation in Gaussian models. *Neural Networks*, *22*(4), 385 – 394.

Pearl, J. (1988). *Probabilistic Reasoning in Intelligent Systems: Networks of Plausible Inference*. Morgan Kaufman Publishers, San Mateo, CA.

Rusmevichientong, P., & Roy, B. V. (2001). An analysis of belief propagation on the turbo decoding graph with Gaussian densities. *IEEE Transactions on Information Theory*, *47*, 745–765.

Seeger, M. W. (2008). Bayesian inference and optimal design for the sparse linear model. *Journal of Machine Learning Research*, *9*, 759–813.

Takahashi, K., Fagan, J., & Chin, M.-S. (1973). Formation of a sparse impedance matrix and its application to short circuit study. In *Proceedings of the 8th PICA Conference*.

Wainwright, M., Jaakkola, T., & Willsky, A. (2003). Tree-reweighted belief propagation algorithms and approximate ML estimation via pseudo-moment matching. In Bishop, C., & Frey, B. (Eds.), *Proceedings of the Ninth International Workshop on Artificial Intelligence and Statistics*. Society for Artificial Intelligence and Statistics.

Watanabe, Y., & Fukumizu, K. (2009). Graph zeta function in the Bethe free energy and loopy belief propagation. In Bengio, Y., Schuurmans, D., Lafferty, J., Williams, C. K. I., & Culotta, A. (Eds.), *Advances in Neural Information Processing Systems 22*, pp. 2017–2025. The MIT Press.

Weiss, Y., & Freeman, W. T. (2001). Correctness of belief propagation in Gaussian graphical models of arbitrary topology. *Neural Computation*, *13*(10), 2173–2200.

Welling, M., & Teh, Y. W. (2001). Belief optimization for binary networks: a stable alternative to loopy belief propagation. In Breese, J. S., & Koller, D. (Eds.), *Proceedings of the 17th Conference in Uncertainty in Artificial Intelligence*, pp. 554–561. Morgan Kaufmann Publishers.

Wiegerinck, W., & Heskes, T. (2003). Fractional belief propagation. In Becker, S., Thrun, S., & Obermayer, K. (Eds.), *Advances in Neural Information Processing Systems 15*, pp. 438–445, Cambridge, MA. The MIT Press.

Yedidia, J. S., Freeman, W. T., & Weiss, Y. (2000). Generalized belief propagation. In *Advances in Neural Information Processing Systems 12*, pp. 689–695, Cambridge, MA. The MIT Press.

Zoeter, O., & Heskes, T. (2005). Change point problems in linear dynamical systems. *Journal of Machine Learning Research*, *6*, 1999–2026.